\documentclass{article}

\usepackage{arxiv}

\usepackage[utf8]{inputenc} % allow utf-8 input
\usepackage[T1]{fontenc}    % use 8-bit T1 fonts
\usepackage{hyperref}       % hyperlinks
\usepackage{url}            % simple URL typesetting
\usepackage{booktabs}       % professional-quality tables
\usepackage{amsfonts}       % blackboard math symbols
\usepackage{nicefrac}       % compact symbols for 1/2, etc.
\usepackage{microtype}      % microtypography
\usepackage{lipsum}
\usepackage{graphicx}
\usepackage{subcaption}
\usepackage{float}

\usepackage{amsmath, amsfonts, amssymb, amsthm, mathtools}

\newtheorem{theorem}{Theorem}[section]

\newtheorem{definition}{Definition}
\newtheorem{example}{Example}

\newcommand{\units}[1]{ \left[ \mathrm{ #1 } \right] }

\newcommand{\R}{ \mathbb{R} }
\newcommand{\Rn}{ \mathbb{R}^n } 
\newcommand{\N}{ \mathbb{N} }
\newcommand{\X}{ \mathcal{X} }
\newcommand{\T}{ \mathcal{T} }
\newcommand{\G}{ \mathcal{G} }
\newcommand{\mat}[1]{ \begin{bmatrix} #1 \end{bmatrix} }

\title{Learning to Estimate Regions of Attraction of Autonomous Dynamical Systems Using Physics-Informed Neural Networks}

\author{
    Cody Scharzenberger \\
    U.S. Naval Research Laboratory
    \and \textbf{Joe Hays} \\
    U.S. Naval Research Laboratory \\
}

\begin{document}

\maketitle

\thispagestyle{fancy}

\begin{abstract}
    When learning to perform motor tasks in a simulated environment, neural networks must be allowed to explore their action space to discover new potentially viable solutions.
    However, in an online learning scenario with physical hardware, this exploration must be constrained by relevant safety considerations in order to avoid damage to the agent’s hardware and environment.
    We aim to address this problem by training a neural network, which we will refer to as a ``safety network'', to estimate the region of attraction (ROA) of a controlled autonomous dynamical system.
    This safety network can thereby be used to quantify the relative safety of proposed control actions and prevent the selection of damaging actions.
    Here we present our development of the safety network by training an artificial neural network (ANN) to represent the ROA of several autonomous dynamical system benchmark problems.
    The training of this network is predicated upon both Lyapunov theory and neural solutions to partial differential equations (PDEs).
    By learning to approximate the viscosity solution to a specially chosen PDE that contains the dynamics of the system of interest, the safety network learns to approximate a particular function, similar to a Lyapunov function, whose zero level set is the boundary of the ROA.
    We train our safety network to solve these PDEs in a semi-supervised manner following a modified version of the Physics Informed Neural Network (PINN) approach, utilizing a loss function that penalizes disagreement with the PDE’s initial and boundary conditions, as well as non-zero residual and variational terms.
    In future work we intend to apply this technique to reinforcement learning agents during motor learning tasks.
\end{abstract}

\section{Introduction}
\label{SEC: Introduction}
Safe and effective control of complex non-linear dynamical systems is a challenging task in the field of robotics.
As robots have begun to take on more difficult motor tasks and to interact increasingly frequently with humans in unstructured environments, there is even greater importance in ensuring that these tools are safe and reliable.
Although traditional control techniques have offered some solutions to this challenge, there has been growing interest in the field of machine learning, and motor learning in particular, as a method for addressing this problem.
This is because neural solutions offer several unique advantages when compared to traditional techniques, including: (1) the ability to adapt to changing system parameters with the introduction of new data, (2) the ability to learn control strategies that are effective across multiple problems, and (3) the possibility of continuing to learn online after deployment.
At the same time, neural solutions to motor control problems raise safety concerns due to their inscrutable nature as ``black box'' solutions.
Equally as important, since most networks for motor learning applications are trained using a reinforcement learning framework, they need to perform exploratory actions to learn how to behave in their environment.
While such exploratory actions are perfectly acceptable and indeed necessary when learning in simulation, they must be performed carefully in online, real world applications to avoid potential damage to robot hardware, property, or life.

To begin addressing this issue, we propose an artificial ``safety network'' that is capable of estimating the complete region of attraction (ROA) of a specific equilibria of a given autonomous dynamical systems.
This work focuses on the development and testing of this safety network, while future work will apply this method to motor learning problems controlled by reinforcement learning agents.
By estimating the ROA of controlled autonomous dynamical systems, this safety network provides two key benefits: (1) it allows for a general evaluation of the stability of the given control scheme, whether this be a neural or traditional controller; and (2) it provides a stability estimate for individual states within the state space of the dynamical system, thereby allowing us to assess the risk of performing exploratory actions near that state.
The safety network accomplishes this task by learning to represent the viscosity solution to a particular partial differential equation (PDE) (see theorem \ref{THEOREM: Yuan-Li Theorem} or \cite{yuan_estimation_2019} for more details), which we refer to as the Yuan-Li PDE, whose solution as $t \to \infty$ is a special function, related to a Lyapunov function, whose zero level set is the complete ROA.
In this way, our work combines existing concepts in Lyapunov stability theory and neural solutions to PDEs to form a novel neural approach for ROA estimation.

\subsection{Related Work}
\label{SEC: Related Work}
Ours is not the first attempt to broach the subject of neural solutions to PDEs, neural ROA estimation, or safe reinforcement learning.
Since \cite{raissi_physics-informed_2019} introduced the Physics Informed Neural Network (PINN) approach to solving PDEs by leveraging automatic differential \cite{griewank_automatic_nodate} to train a fully connected ANN endowed with a loss function that encodes information about the underlying PDE, many other studies have sought to apply, extend, or add theoretical backing to this method.
\cite{shin_convergence_2020} has studied the efficacy of PINNs for generating solutions to linear elliptic and parabolic PDEs, while \cite{cai_physics-informed_2021} and \cite{belbute-peres_combining_2020} apply the PINN framework to generate heat-diffusion and fluid flow solutions, respectively. 
\cite{kharazmi_variational_2019,khodayi-mehr_varnet_2019,kharazmi_hp-vpinns_2021} all attempt to expand the PINN framework to include loss terms that penalize non-zero PDE variation, which provides information about the PDE over areas of the input domain, not just at specific collocation points.
Others have sought to extend PINNs to be applicable to a wider variety of PDE problems, including PDEs with fractional powers \cite{pang_fpinns_2019} or where solutions to parameterized PDEs are desired \cite{dal_santo_data_2020}.
At the same time, efforts have been made to improve convergence rates with adaptive activation functions \cite{jagtap_locally_2020,jagtap_adaptive_2020,mcclenny_self-adaptive_2020}, intelligent selection of loss weightings \cite{wang_understanding_2020}, or subdomain approaches that use multiple PINNs across the domain of interest stitched together with appropriate interface conditions \cite{jagtap_conservative_2020,karniadakis_extended_2020}.
Related efforts have also attempted to quantify the generalization error of these methods for specific problems \cite{shin_convergence_2020,yang_adversarial_2019,zhang_quantifying_2019,mishra_estimates_2021} and explain the convergence difficulties that PINNs face when solving some PDEs \cite{wang_understanding_2020}.
Although much recent effort has been dedicated to investigating the PINN framework for solving PDEs, other neural PDE solution techniques exist, notably techniques which learn iterators that are not solutions to PDEs themselves but rather provide a method of quickly computing such solutions \cite{hsieh_learning_2019}.

While the recently developing field of scientific machine learning (SciML) has made quick progress on the subject of neural solutions to PDEs, attempts at machine learning based ROA estimation and safe reinforcement learning have mostly been disconnected from this research.
\cite{berkenkamp_safe_2016-1} suggests a Gaussian process (GP) \cite{neal_regression_nodate} approach to learning ROAs for physical systems from experimental data, while \cite{berkenkamp_safe_2015,berkenkamp_safe_2016} uses a similar GP model to safety tune the parameters of a controller while maintaining stability.
Along the same lines, \cite{berkenkamp_safe_2017} combines Bayesian inference \cite{dempster_generalization_1968} and GPs to simultaneous learn both the ROA of some dynamical system and an optimal control scheme while maintaining provably safe stability certificates.
On the subject of neural ROA estimation, \cite{richards_lyapunov_nodate} uses a supervised learning framework to teach a fully connected ANN with a modified output architecture to approximate a Lyapunov function whose unity level set approaches the complete ROA.

\subsection{Our Contribution}
\label{SEC: Our Contribution}
The work presented herein uniquely combines the fields of neural solutions to PDEs with Lyapunov stability theory to create the novel safety network framework for estimating ROAs for autonomous dynamical systems with modified PINNs.
While several of the aforementioned recent works estimate ROAs for dynamical systems, they do so either in a Bayesian framework \cite{berkenkamp_safe_2016-1} or in a supervised learning framework that does not leverage information about the underlying ROA dynamics \cite{richards_lyapunov_nodate}.
Instead of learning a single Lyapunov function whose unity level set is the ROA for the dynamical system of interest, our safety network learns to evolve an initial ROA estimate forward in time to a final ROA estimate as governed by the Yuan-Li PDE that implicitly encodes ROA estimates as the zero level sets of its solution at each time step.
Compared to \cite{berkenkamp_safe_2016-1}, our safety network method is a neural ROA estimation technique that is inextricably linked to the solution of a PDE that describes ROA dynamics.
This poses several unique advantages, including: (1) incorporating prior knowledge about ROA dynamics inherited from the Yuan-LI PDE; (2) learning to evolve initial ROA estimates into complete ROAs, assisting in generalizing between related problems; and (3) the possibility of learning to solve ROAs for whole families of dynamical systems through parameterized PDE solutions techniques (not investigated here, but a promising subject of future study).
Compared to numerical ROA estimation techniques, our safety network method has the typical strengths normally attributed to neural solutions, such as greater generalizability and the ability to adapt to new data, but sacrifices the thorough theoretical foundations of numerical methods for our still nascent theoretical understanding of PINNs. 
While the dependence of our method on neural PDE solutions techniques allows it to improve as our understanding of scientific machine learning grows, it is not without a few notable limitations.
Firstly, the Yuan-Li PDE only describes the ROA dynamics of autonomous dynamical systems, which represents a useful but not all encompassing subset of dynamical systems.
Secondly, it can be quite difficult to generate solutions to the Yuan-Li PDE for high dimensional (curse of dimensionality) or highly non-linear systems with complex dynamics without extensive hyperparameter tuning (see section \ref{SEC: Discussion}).
As we shall see, some of these difficulties are due to the underlying PINN approach, while others are inherent to the Yuan-Li PDE itself.
Considering both the advantages and limitations of the safety network method, we see this approach as a promising and useful alternative for neural ROA estimation that successfully leverages our understanding of the development of ROAs for autonomous dynamical systems.

\subsection{Paper Organization}
\label{SEC: Paper Organization}
In this paper, we begin our analysis in section \ref{SEC: Background} by introducing the key components of Lyapunov stability theory, numerical solutions to PDEs, and neural solutions to PDEs that form the foundation of this work.
We then fully develop the methodology that we use to train our safety networks to predict the ROAs of dynamical systems in section \ref{SEC: Methods}, focusing on the loss function that we minimize during training.
Next, we validate the efficacy of our method by demonstrating its ability to predict the complete ROA of several example autonomous dynamical systems in section \ref{SEC: Results}.
Through these examples we show both that our safety network is able to learn ROAs of various shapes and dimensions, as well as the ability of the network to quickly learn the solution to new related problems via pre-training.
Finally, we discuss the implications of these results in section \ref{SEC: Discussion}, the future work that we intend to perform in this field in section \ref{SEC: Future Work}, and the conclusions we have drawn from this work in section \ref{SEC: Conclusion}.

\section{Background}
\label{SEC: Background}
A central component of this work is the approximation of regions of attraction (ROAs) for autonomous dynamical systems using ANNs.
As we shall see, this process requires the solution to the Yuan-Li PDE, which necessitates the synthesis of methods from across several disciplines, including Lyapunov stability theory, numerical solutions to PDEs, and neural solutions to PDEs.
In this section, we introduce the fundamental concepts of the prior work in each of these fields that are necessary to understand how we train and evaluate an ANN that is capable of approximating the ROA of specific autonomous dynamical systems.
Some of the mathematical formulations presented differ in appearance from their source material because they have been superficially altered to conform to the standardized notational scheme used throughout this work.
See Table \ref{TABLE: Variable Definitions} in Appendix \ref{SEC: Table of Variables} for a description of the variables used throughout this work.

\subsection{Lyapunov Stability Theory}
\label{SEC: Lyapunov Stability Theory}
While linear dynamical systems have single equilibria that are either globally stable, marginally stable, or unstable, non-linear dynamical systems tend to exhibit much more complicated behavior.
It is possible for non-linear dynamical systems to have any number of equilibria and for the stability of these equilibria to be restricted to local regions in the state space \cite{haddad_nonlinear_2011}.
In order to adequately describe these complexities, Lyapunov stability theory provides several definitions that we will refer to throughout this work.

\begin{definition}[Lyapunov Function]
    Given an autonomous dynamical system $\dot{x} = f(x)$ with $f: \X \subset \Rn \to \Rn$ and an equilibrium point $x_e \in \X^o$, a strict Lyapunov function is a continuously differentiable, positive definite, scalar function $u: \X \to \R$ with negative definite time derivative, $\dot{u} = \nabla u(x) \cdot f(x)$.
    A non-strict Lyapunov function is identical, except it permits a semi-negative definite time derivative.
\end{definition}

Lyapunov functions are important in several ways, including how they relate to stability and ROAs.
Every dynamical system will have one, many, or no Lyapunov functions associated with each of its equilibrium points.
If there are no Lyapunov functions associated with an equilibrium point, then this equilibrium point is unstable.
On the other hand, if there are one or more Lyapunov functions associated with an equilibrium point, then this equilibrium point is stable.
For stable equilibrium points, one further distinction is made: if at least one of the associated Lyapunov functions is strict, then the equilibrium point is said to be asymptotically stable; if not, the equilibrium point is simply said to be (Lyapunov) stable.
See \cite{haddad_nonlinear_2011} for a more in depth discussion of these concepts.

Lyapunov functions not only provide information about the stability of an equilibrium point, but also about the ROA surrounding stable equilibrium points.

\begin{definition}[Region of Attraction (ROA)]
    Given an autonomous dynamical system $\dot{x} = f(x)$ with $f: \X \subset \Rn \to \Rn$ and an equilibrium point $x_e \in \X^o$, the associated complete region of attraction (ROA) is defined as
    
    \begin{align}
        \mathcal{R}_{x_e} = \left\{ x_0 \in \X : \lim_{t \to \infty} \psi_{x_0}(t) = x_e \right\}
    \end{align}

    \noindent where $\psi_{x_0}: [0, \infty) \to \Rn$ is the trajectory of the dynamical system with initial condition $\psi(0) = x_0$.
    Any subset of the complete ROA is an incomplete ROA.
\end{definition}

\noindent Given this definition, the complete ROA of a stable equilibrium point is the set of all initial states from which the flow of the dynamical system converges to the equilibrium point.
The complete ROA of an unstable equilibrium point is simply the equilibrium point itself, while the complete ROA of a stable equilibrium may have some arbitrary shape that depends on the local flow field.
The descriptors ``complete'' and ``incomplete'' are often dropped when the distinction is irrelevant or clear from the context.

Each Lyapunov function $u_{x_e}: \X \subset \Rn \to \R$ associated with an equilibrium point $x_e \in \X^o$ yields a ROA $\mathcal{R}_{u_{x_e}} \subset \mathcal{R}_{x_e}$. 
We write $\mathcal{R}_{x_e}$ to indicate the complete ROA associated with an equilibrium point $x_e \in \X^o$ and $\mathcal{R}_{u_{x_e}}$ to indicate the ROA associated with a specific (Lyapunov) function $u$ (which may or may not be the complete ROA) for the equilibrium point $x_e$.
For each equilibrium point $x_e \in \X^o$, there exists a unique Lyapunov function $u_{x_e}: \X \subset \Rn \to \R$ whose associated ROA $\mathcal{R}_{u_{x_e}}$ is the complete ROA.
Zubov's theorem provides a method through which this unique Lyapunov function may be determined \cite{margolis_control_1963}.

\begin{theorem}[Zubov's Theorem]
    \label{THEOREM: Zubov's Theorem}
    Given an autonomous dynamical system $\dot{x} = f(x)$ with $f: \X \subset \Rn \to \R$ and an equilibrium point $0 \in \X^o$, a set $\mathcal{R} \subset \X$ with $0 \in \mathcal{R}^o$ is the ROA of $0$ iff $\exists u, h$ such that:
    \begin{itemize}
        \item $u(0) = h(0) = 0$, $0 < u(x) < 1$ for $x \in \mathcal{R} \setminus \{0\}$, and $h > 0$ for $x \in \X \setminus \{0\}$.
        \item $\forall \gamma_2 > 0$, $\exists \gamma_1 > 0$ and $\alpha_1 > 0$ such that $u(x) > \gamma_1$ and $h(x) > \alpha_1$ if $||x|| > \gamma_2$.
        \item $u(x_n) \to 1$ for $x_n \to \partial \mathcal{R}$ (or $||x_n|| \to \infty$).
        \item $\nabla u(x) \cdot f(x) = - h(x) (1 - u(x)) \sqrt{1 + ||f(x)||^2}$
    \end{itemize}
\end{theorem}

\noindent While the statement of Zubov's theorem has several parts, the central point is that the solution to Zubov's PDE yields a Lyapunov function $u_{x_e}: \X \subset \Rn \to \R$ whose associated ROA $\mathcal{R}_{u_{x_e}}$ is the complete ROA $\mathcal{R}_{x_e}$ (not just a conservative estimate of the complete ROA).
Written succinctly, if $u_{x_e}: \X \subset \Rn \to \R$ satisfies Zubov's PDE, then $\mathcal{R}_{u_{x_e}} = \mathcal{R}_{x_e}$.
Unfortunately, like many complex PDEs, Zubov's PDE does not in general have an analytical solution.
Nonetheless, several different numerical approaches have been developed to compute an approximate solution to Zubov's PDE, and exact solutions have been developed for specific applications \cite{rezaiee-pajand_estimating_2012}.

Zubov's PDE is not the only way to determine an implicit, level set based representation of the complete ROA associated with an equilibrium point.
Yuan and Li show that the viscosity solution to a particular Hamilton-Jacobi PDE yields a special function $u_{x_e}: \X \subset \Rn \to \R$ whose zero level set $\partial \mathcal{L}_{0, u_{x_e}} = \{ x \in \X : u_{x_e}(x) = 0 \}$ is the boundary of the complete ROA (i.e., $\partial \mathcal{L}_{0, u_{x_e}} = \partial \mathcal{R}_{x_e}$).
While the function generated by solving the Yuan-Li PDE as $t \to \infty$ is not a Lyapunov function, it is related to the backward reachable set of the dynamical system, and therefore can serve as an implicit representation of the ROA just like a Lyapunov function (see \cite{yuan_estimation_2019} for more details).
Note that we use the notation $\partial \mathcal{L}_{0, u_{x_e}} = \{ x \in \X : u_{x_e}(x) = 0 \}$ to denote the zero level set of the function $u_{x_e}: \X \subset \Rn \to \R$ and the notation $\mathcal{L}_{0, u_{x_e}} = \{ x \in \X : u_{x_e}(x) \leq 0 \}$ to denote the zero sublevel set of the same function.

\begin{theorem}[Yuan-Li Theorem]
    \label{THEOREM: Yuan-Li Theorem}
    Let $\X \subset \Rn$ and $\T = (-\infty, 0]$.
    Consider an autonomous dynamical system $\dot{x} = f(x)$ with $f(x): \X \to \Rn$ and a stable equilibrium point $x_e \in \X^o$.
    Suppose that $\phi(x, t): \X \bigtimes \T \to \R$ is the viscosity solution of the terminal value Hamilton-Jacobi equation

    \begin{align}
        \frac{\partial \phi}{\partial t} + \min \left[ 0, \left( \frac{\partial \phi}{\partial x} \right)^T f(x) \right] &= 0 \\
        \phi(x, 0) &= \phi_0(x)
    \end{align}

    \noindent where $\phi_0$ is bounded and Lipschitz continuous, and $x_e \in \mathcal{L}_{0, \phi_0} \subset \mathcal{R}_{x_e}$.
    Let the function $u_{x_e}: \X \to \R$ be defined as
    
    \begin{align}
        u_{x_e}(x) = \lim_{t \to -\infty} \phi(x, t).        
    \end{align}
    
    \noindent Then the complete ROA associated with $x_e$ is 

    \begin{align}
        \mathcal{R}_{x_e} = \mathcal{L}_{0, u_{x_e}}.
    \end{align}
\end{theorem}

\noindent As with Zubov's theorem, Yuan and Li's theorem relates the ROA of a stable equilibrium point to the solution of a particular PDE.
In this case, the procedure is to solve the Yuan-Li PDE for $\phi(x, t): \X \bigtimes \T \to \R$, then to compute the associated function $u_{x_e}(x) = \lim_{t \to -\infty} \phi(x, t)$, and finally to compute the zero sublevel set $\mathcal{L}_{0, u_{x_e}}$ that gives the complete ROA $\mathcal{R}_{x_e}$.
Since the Yuan-Li PDE is simpler to solve than Zubov's PDE and does not require auxiliary functions, we opt to solve their PDE to determine ROAs.
Zubov's PDE could equally well be used as the basis for training our safety network, but for the purpose of this work is left as key historical context.

\subsection{Numerical ROA Solutions}
\label{SEC: Numerical ROA Solutions}
Per the above discussion of Lyapunov stability theory, it is clear that estimating the ROAs of autonomous dynamical systems can be accomplished by computing the solution of specially chosen PDEs.
Unfortunately, the complexity of these PDEs means that analytical solutions are only available in limited circumstances and are not widely applicable.
It is therefore common to generate ROA approximations using appropriate PDE numerical solution techniques.
Note that we do not depend on these numerical solutions to train our safety network.
Instead, we mention numerical methods here because: (1) we use them to compute high fidelity numerical ROA estimates for the sake of comparison with our neural approach; and (2) they demonstrate a key difficulty in solving the Yuan-Li PDE, namely that the PDE solution tends to be highly discontinuous near the ROA boundary.
To overcome the challenge that such discontinuity poses to many numerical methods, we use the same techniques that Yuan and Li propose for solving Hamilton-Jacobi PDEs of the kind used in their theorem \cite{yuan_estimation_2019}.
Specifically, we use a fifth order weighted essentially non-oscillatory (WENO5) scheme to discretize the spatial variables of the PDE \cite{liu_weighted_1994} and then apply a third order total variation decreasing Runge-Kutta (TVDRK3) method to integrate the system through time \cite{gottlieb_total_1998}.
This approach to solving PDEs is sometimes referred to as the method of lines \cite{sadiku_simple_2000}, because a spatial discretization scheme is used to convert the PDE into a system of ODEs that can be integrated with ODE solvers.
For some detail regarding our implementation of these numerical methods, see Appendix \ref{SEC: Numerical Methods} and the aforementioned source material.

\subsection{Neural Solutions to PDEs}
\label{SEC: Neural Solutions to PDEs}
As the field of machine learning has matured, there has been a growing interest in applying the techniques of machine learning to scientific computing, resulting in the rapid development of the field of scientific machine learning (SciML).
An emerging area of study in this field is the solution of PDEs using neural networks.
Interest in this topic stems from the powerful expressivity and generalizability of neural networks coupled with the fact that PDEs are used to model many natural phenomenon across many scientific and engineering disciplines.
If neural networks were able to represent the solutions to some of these fundamental mathematical models, they would not only become a tool for investigating challenging scientific problems but would also bring several of their unique advantages to these fields of study.
Compared to the traditional numerical methods used to solve PDEs, neural networks offer the possibility of adapting their solutions over time as new data is introduced.
Similarly, with advances in meta-learning and other techniques, it may become possible for neural networks to learn to represent the solutions to whole families of related PDEs, not just a single specifically chosen PDE \cite{dal_santo_data_2020}.  
While these interesting applications are the subject of future studies, they both rely on the fundamental ability of neural networks to learn to represent the solutions to PDEs, which is a central component of this work.

There are several different approaches that can be taken when training an ANN to approximate the solution of a PDE.
If an exact analytical solution is known, the network can be trained within a supervised learning framework to represent the known solution.
This is possible even with a single hidden layer given a sufficiently large layer width due to the well known fact that such networks are universal function approximators \cite{hornik_multilayer_1989}.
While effective, this method is of limited practical utility because it is only applicable to problems for which exact or high fidelity numerical solutions are known apriori.
If such a solution is not available, one can instead train the network to represent the solution to a PDE using any of a number of semi-supervised learning strategies based on the PINN framework \cite{raissi_physics-informed_2019}.
This is done by enforcing agreement with any relevant initial and boundary conditions while simultaneously ensuring that the network satisfies the governing differential equation.
Since initial condition (IC) and boundary condition (BC) information must be known to generate an exact solution to any PDE, a network can always be trained to satisfy the ICs and BCs of a PDE by including terms in the loss function that penalize disagreement between the network predictions and these conditions.
At the same time, in order to ensure that the network satisfies the governing differential equation, information about the PDE itself must be incorporated into the loss function when using this framework.
Although there are several different proposed strategies (see sources in section \ref{SEC: Related Work}), we accomplish this in our safety network by incorporating loss terms that penalize non-zero PDE residual and variation.

\begin{definition}[Yuan-Li PDE Residual]
    Let $\X \subset \Rn$ and $\T = (-\infty, 0]$.
    Given an autonomous dynamical system $\dot{x} = f(x)$ with $f: \X \to \Rn$ and the associated Yuan-Li PDE
    
    \begin{align}
        \dot{\phi} = - \min \left( 0, \nabla \phi^T f(x) \right)
    \end{align}
    
    \noindent with $\phi(x, t): \X \bigtimes \T \to \R$, the PDE residual $r(x, t): \X \bigtimes \T \to \R$ is

    \begin{align}
        r(x, t) = \dot{\phi} + \min \left( 0, \nabla \phi^T f(x) \right),
    \end{align}
\end{definition}

\noindent As a direct consequence of this definition, $r(x, t) = 0$ $\forall x \in \X$ and $\forall t \in \T$.
This means that the exact solution to the PDE will have zero residual everywhere in the spatiotemporal domain of the problem, giving us our first method of evaluating the extent to which our neural network solution agrees with the underlying differential equation.

Thus far we have distinguished between the spatial and temporal variables of PDEs by denoting the spatial variables as $x \in \X$ and the temporal variables as $t \in \T$.
We will continue to use this notation when the distinction between the spatial and temporal variables is relevant.
However, it is often useful to denote the entire vector of spatiotemporal states as a single variable, in which case we adopt the notation $s \in \mathcal{S} = \X \bigtimes \T$ where $s = \mat{ x & t }^T$.
Using this notation, we can succinctly define the PDE variation.

\begin{definition}[PDE Variation]
    Let $\X \subset \Rn$, $\T \subset \R$, and $\mathcal{S} = \X \bigtimes \T$.  For any compactly supported function $g: \mathcal{S} \to \R$ and PDE residual $r: \mathcal{S} \to \R$, the associated PDE variation $v_g \in \R$ is defined by

    \begin{align}
        v_g = \int_{\mathcal{S}} g(s) r(s) ds.
    \end{align}
\end{definition}

\noindent Note that the above variational form is a definite integral over the spatiotemporal domain $\mathcal{S}$ and therefore yields a scalar $v_g \in \R$ for each $g: \mathcal{S} \to \R$.
Similarly, since $r(s) = 0$ $\forall s \in \mathcal{S}$, we know that $v_g = 0$ for every compactly supported function $g: \mathcal{S} \to \R$.
As before, this means that the exact solution to the PDE will have zero variation over the spatiotemporal domain for every compactly supported auxiliary function, providing a second method of evaluating the extent to which our neural network solution agrees with the underlying differential equation.
The specific auxiliary functions that are most effective for training networks of this type is still an open area of research.
\cite{kharazmi_variational_2019} and \cite{khodayi-mehr_varnet_2019} use polynomial functions, while \cite{kharazmi_hp-vpinns_2021} compares several options.

Regardless of the choice of auxiliary function, we need to be able to evaluate the associated integral in order to use the PDE variation in a loss function for training.
While there are multiple approaches that could be taken to address this problem, we opt to simplify the integral by choosing our compactly supported auxiliary functions $g$ to be the linear basis functions of cuboid finite elements, a set that we call $\G$.
We then integrate these basis functions over $\mathcal{S}$ via Gauss-Legendre quadrature, using the same procedure as \cite{khodayi-mehr_varnet_2019}.
These basis functions and quadrature method are described in the following sections.

\subsubsection{Gauss-Legendre Quadrature (GLQ)}
\label{SEC: Gauss-Legendre Quadrature}
Gauss-Legendre quadrature (GLQ) provides a simple and fast method of approximating the integral of scalar functions over the unit interval \cite{swarztrauber_computing_2003}.
Despite being most commonly written in a single dimension, GLQ is applicable to integrals of any dimension with the appropriate extensions.
Since we will need to integrate functions of arbitrary dimension, we will begin with the one-dimensional GLQ rule and then extend the method to arbitrary dimension.

\begin{theorem}[1D Gauss-Legendre Quadrature]
    Consider a scalar function $h: \R \to \R$ and a domain $D = [-1, 1]$.  The integral of $h$ over $D$ can be approximated by

    \begin{align}
        \int_{-1}^{1} h(x) dx \approx \sum_{i = 1}^{n} w_i h(x_i)
    \end{align}

    \noindent with

    \begin{align}
        w_i = \frac{2}{(1 - x_i^2)(P_n'(x_i))^2}
    \end{align}

    \noindent where $n$ is the number of quadrature points, $w_i$ are the quadrature weights, $x_i$ are the quadrature roots (specifically selected to be the roots of the $n$th order Legendre Polynomial), $P_n$ is the $n$th order Legendre Polynomial, and $P_n'$ is the derivative of the $n$th order Legendre Polynomial.
\end{theorem}

\noindent To apply this method, we first select a number of quadrature points $n$, then compute the $n$th order Legendre Polynomial $P_n$ along with its derivative $P_n'$ and roots $x_i$, and finally use this information to compute the quadrature weights and approximate the integral.

In order to extend this method to higher dimensions, we approximate the integral of each dimension using the one-dimensions GLQ rule and then perform a change of indexes to put the result into a recognizable form.
Consider a multi-input scalar-valued function $h: \R^{d} \to \R$ and domain $D = [-1, 1]$.
Let $x = (x_1, \dots, x_d) \in \R^d$ be an arbitrary point in the domain of $h$ such that we can write $h(x_1, \dots, x_d)$ to emphasize that $h$ is a function of $x_1, \dots, x_d \in \R$.
Also, let $\Xi \subset \R^d$ with $\Xi = D \bigtimes \cdots \bigtimes D$ ($d$ Cartesian products of $D$) so that $\Xi$ is the $d$-dimensional unit hyper-cube.
Then the integral of $h$ over the volume $\Xi$ can be approximated by

\begin{align}
    \int_{\Xi} h(x) dx = \int_{-1}^{1} \cdots \int_{-1}^{1} h(x_1, \dots, x_d) dx_1 \cdots dx_d \approx \sum_{i_d = 1}^{n_d} \cdots \sum_{i_1 = 1}^{n_1} w_{d, i_d} \cdots w_{1, i_1} h(x_1, \dots, x_d)
\end{align}

\noindent Let $x_k = \mat{ x_{1, k} & \cdots & x_{d, k} }^T$ where $k \in \{ 1, \dots, \left( n_1 \cdots n_d \right) \}$ defined by $k = \sum_{j = 1}^d (i_j - I_j) \prod_{l = 1}^{j - 1} n_{l}$ with $I_{1} = 0$ and $I_{j \neq 1} = 1$.
Under this definition, $k$ is a new index that is computed from the existing indexes $i_j$.
Finally, let $w_k = \prod_{j = 1}^{d} w_{j, i_j}$.
Then we can condense the above result to

\begin{align}
    \int_{\Xi} h(x) dx \approx \sum_{i_d = 1}^{n_d} \cdots \sum_{i_1 = 1}^{n_1} w_{d, i_d} \cdots w_{1, i_1} h(x_1, \dots, x_d) \approx \sum_{k = 1}^{n_{1} \cdots n_{d}} w_k h(x_k).
\end{align}

\noindent Notice that the final quadrature rule in every dimension has the same generic form with different indexes. 

\subsubsection{Finite Element Basis Functions}
\label{SEC: Finite Element Basis Functions}
There are many different basis functions that can be used to interpolate scalar quantities over finite elements of arbitrary shape and dimension.
To improve the computational plausibility of our method and simplify the following analysis, we consider only cuboid linear elements in this work, although we acknowledge that there are other valid choices that could be made.
For the purpose of this discussion, let $d \in \N$ be the dimension of the finite element in question.
A one dimensional (i.e., $d = 1$) cuboid linear element is simply an interval with two basis functions (one associated with each end of the interval).
These basis functions can be computed by linearly interpolating between the points $(x_1, y_1) \in \R^{2}$ and $(x_2, y_2) \in \R^{2}$, where $x_1, x_2 \in \R$ are the end points of the interval in question and $y_1, y_2 \in \R$ are the known scalar values at these end points.
The line connecting these end points satisfies both $y_1 = m x_1 + b$ and $y_2 = m x_2 + b$, where $m, b \in \R$ represent the slope and y-intercept of the line, respectively.
Solving for $m$ and $b$, we have

\begin{align}
    m = \frac{y_2 - y_1}{x_2 - x_1} \,\,\, \& \,\,\, b = \frac{y_1 x_2 - y_2 x_1}{x_2 - x_1}
\end{align}

\noindent \\ Substituting this slope and $y$-intercept into our line equation yields

\begin{align}
    y &= \left( \frac{y_2 - y_1}{x_2 - x_1} \right) x + \left( \frac{y_1 x_2 - y_2 x_1}{x_2 - x_1} \right) = \left( \frac{x_2 - x}{x_2 - x_1} \right) y_1 + \left( \frac{x - x_1}{x_2 - x_1} \right) y_2
\end{align}

\noindent \\ Looking at the right-hand expression, we see that the two 1D linear basis functions are

\begin{align}
    g_1(x) = \frac{x_2 - x}{x_2 - x_1} \,\,\, \& \,\,\, g_2(x) = \frac{x - x_1}{x_2 - x_1}
\end{align}

\noindent \\ If we assume that we are defining basis functions on the unit interval (as we will be when we integrate via GLQ), the basis functions become

\begin{align}
    g_{1}(x) = 0.5 (1 - x) \,\,\, \& \,\,\, g_{2}(x) = 0.5 (1 + x).
\end{align}

As before, we are actually interested in the basis functions for finite elements of arbitrary dimension.
Fortunately, we can construct the higher dimensional basis functions by taking the appropriate products of the one-dimensional basis functions.
As noted above, each element vertex has its own basis function, meaning that for a $d$-dimensional linear element, we have $2^d$ basis functions.
Let $i_j \in \{ 1, 2 \}$ $\forall j \in \{1, \cdots, d\}$ be indexes that describe the location of each vertex on the linear element with respect to the $j$th dimension.
Then let $(i_1, \dots, i_d)$ be the index for the vertex of interest.
The basis function associated with this vertex is

\begin{align}
    g_{i_1, \dots, i_d}(x) = g_{i_1}(x_1) \cdots g_{i_d}(x_d)
\end{align}

\section{Methods}
\label{SEC: Methods}
With the requisite mathematical foundations established, we now synthesize these techniques into a single method to train our safety network to approximate the ROA of several example autonomous dynamical systems.
We begin by discussing the safety network architecture and hyperparameter selection; next we detail the loss function that we minimize to get the desired network behavior; then we address training data generation, minibatching, and initial conditions; and finally, we discuss the experiments that we perform using our network to validate its efficacy.

\subsection{Network Architecture \& Hyperparameters}
\begin{figure}[htbp]
	\centering
  	\includegraphics[width=4.5in]{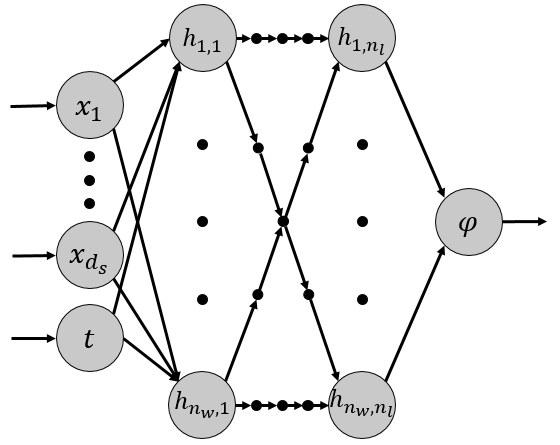}
  	\caption{Fully connected feed-forward safety network architecture.}
    \label{FIG: NetworkArchitecture}
\end{figure}

Our safety network uses a simple fully connected feed-forward architecture whose weights and biases are determined by applying the Adam optimizer \cite{kingma_adam_2017} to a specific loss function (see Section \ref{SEC: Loss Function}).
This network has a number of inputs equal to the total spatiotemporal dimension of the dynamical system being investigated, a single output neuron that provides an estimate of the stability of the given state, and $n_{layers} = 3$ hidden layers each with $n_{width} = 50$ neurons using the hyperbolic tangent $\tanh()$ activation function.
The architecture in question is depicted in Fig. \ref{FIG: NetworkArchitecture}.

\noindent Note that depending on the complexity and dimension of the underlying dynamical system whose ROA we are approximating, it may be necessary to augment this basic network structure with more layers and neurons per layer to ensure that the network has the expressive power necessary to capture the potentially complex shape of the ROA.
Other hyperparameters that we must select for our method include the learning rate, GLQ order, and loss weights.
We use a fixed learning rate of $0.005$, a GLQ order of $n_o = 1$, and loss weights of $c_{IC} = 1$, $c_{BC} = 0.1$, $c_{Mon} = 10$, $c_{r} = 1$, $c_{v} = 1$, and $c_{reg} = 1e-5$.
Note that the selection of appropriate loss weights can be a difficult problem that varies depending on the PDE.
As such, some have suggested automatic methods of setting these parameters \cite{wang_understanding_2020}.
Next we take an in-depth look at the loss function we use to learn network weights and biases.

\subsection{Loss Function}
\label{SEC: Loss Function}
Consider an autonomous dynamical system $\dot{x} = f(x)$ with $f: \X \subset \R^{d_s} \to \R^{d_s}$, $d_s \in \N$ spatial dimensions, and a single temporal dimension $d_t = 1$.
Let the spatial domain of this dynamical system be $\X \subset \R^{d_s}$ and the temporal domain be $\T \subset \R^{d_t}$ defined by $\T = [0, t_{max}]$ where $t_{max} \in \R_{+}$.
The total number of spatiotemporal dimensions is the sum of these constituent dimensions, namely $d = d_s + d_t = d_s + 1$, and the complete spatiotemporal domain is $\mathcal{S} = \X \bigtimes \T \subset \R^{d}$ where the $s \in \mathcal{S}$ have the form $s = \mat{ x & t }^T$.
Suppose that the dynamical system has a stable equilibrium point $x_e \in \X^o$ whose complete ROA $\mathcal{R}_{x_e}$ we wish to approximate via an ANN with the aforementioned architecture.
Without loss of generality, we can assume that $x_e = 0$, because in the case that we are interested in a non-zero equilibrium point, we can simply shift the flow field $f(x)$ such that the desired equilibrium point is at the origin.

As shown by Yuan and Li, the complete ROA $\mathcal{R}_{x_e}$ for a dynamical system with flow $f: \X \to \R^d$ and equilibrium point $x_e \in \X^o$ is the zero sublevel set $\mathcal{L}_{0, u_{x_e}}$ of the function $u_{x_e}: \X \to \R$ defined by $u_{x_e}(x) = \lim_{t \to \infty} \phi(x, t)$, where $\phi: \mathcal{S} \to \R$ is the solution to the PDE

\begin{align}
    \dot{\phi} = - \min \left( 0, - \nabla \phi^T f(x) \right)
\end{align}

\noindent with initial condition $\phi_0(x) = \phi(x, 0)$, provided that $\phi_0(x)$ is bounded and Lipschitz continuous, and that $x_e \in \mathcal{L}_{0, \phi_0} \subset \mathcal{R}_{x_e}$.
Therefore, if we can train an ANN to approximate the function $\phi$, we can compute the associated function $u_{x_e}$ and use its zero sublevel set $\mathcal{L}_{0, u_{x_e}}$ to approximate the complete ROA $\mathcal{R}_{x_e}$.
Let our network approximation of $\phi$ be $\phi^{\star}$, the associated function be $u_{x_e}^{\star}$, and the estimated ROA be $\mathcal{R}_{x_e}^{\star}$.
In order to get the network's approximation $\phi^{\star}$ as close as possible to the true solution $\phi$, we train the network in a semi-supervised manner by utilizing a loss function comprised of a linear combination of terms that penalize: (1) disagreement with PDE ICs and BCs, (2) non-monotonicity, (3) non-zero PDE residual, (4) non-zero PDE variation, and (5) weight and bias magnitude.
We refer to these loss terms as the IC loss, BC loss, monotonicity loss, residual loss, variational loss, regularization losses, respectively.

\subsubsection{Initial Condition Loss}
\label{SEC: Initial Condition Loss}
The initial condition loss is

\begin{align}
    L_{IC} = \sqrt{\sum_{i = 1}^{n_{IC}} \left( \phi(s_{IC, i}) - \phi^{\star}(s_{IC, i}) \right)^2},
\end{align}

\noindent where $s_{IC, i} \in \mathcal{S}_{IC} = \X \bigtimes \{0\} \subset \mathcal{S}$ and $n_{IC} \in \N$ is the total number of initial condition training points.
To clarify, each $s_{IC, i} \in \mathcal{S}_{IC}$ has the form $s_{IC, i} = \mat{ x_i & 0 }^T$ where $x_i \in \X$.
The IC loss is simply the least squares error of the network prediction $\phi^{\star}$ evaluated at each initial condition point.
As is evident in the name and its formulation, the purpose of the IC loss is to encourage agreement of the network's approximation $\phi^{\star}$ with the specified initial conditions.

\subsubsection{Boundary Condition Loss}
\label{SEC: Boundary Condition Loss}
Similarly, the boundary condition loss is

\begin{align}
    L_{BC} = \sqrt{\sum_{i = 1}^{n_{BC}} \left( \phi(s_{BC, i}) - \phi^{\star}(s_{BC, i}) \right)^2},
\end{align}

\noindent where $s_{BC, i} \in \mathcal{S}_{BC} = \partial \X \bigtimes \T \subset \mathcal{S}$ and $n_{BC} \in \N$ is the total number of boundary condition training points.
To clarify, each $s_{BC, i} \in \mathcal{S}_{BC}$ has the form $s_{BC, i} = \mat{ x_{BC, i} & t_i }^T$ where $x_{BC, i} \in \partial \X$ and $t_i \in \T$.
This means that the BC loss is simply the least squares error of the network prediction $\phi^{\star}$ evaluated at each boundary condition point.
Just as the IC loss helped to encourage agreement between the network's approximation $\phi^{\star}$ and the specified initial condition, so does the BC loss encourage agreement between the network's approximation $\phi^{\star}$ and the specified boundary condition.

\subsubsection{Monotonicity Loss}
\label{SEC: Monotonicity Loss}
In order to ensure that the ANN ROA estimate $\mathcal{R}_{x_e}^{\star}$ grows over time, the ANN viscosity solution estimate $\phi^{\star}$ must strictly decrease over time.
While this behavior is encoded in the construction of the Yuan-Li PDE, it is strictly enforced when using numerical methods by only permitting updates of the appropriate sign.
Our experiments indicate that incorporating a similar penalty for increasing $\phi^{\star}$ estimates improves convergence to a correct result.
We call this loss the monotonicity loss and define it as

\begin{align}
    L_{Mon} = \sqrt{\sum_{i = 1}^{n_{c}} \left( \min \left( \phi(x_{c, i}, 0) - \phi^{\star}(x_{c, i}, t_{c, i}), 0 \right) \right)^2},
\end{align}

\noindent where $x_{c, i} \in \X$ are the spatial components of the collocation points, $t_{c, i} \in \T$ are the temporal components of the collocation points, and $n_c \in \N$ is the number of collocation points.
This loss term does not strictly enforce the requirement that $\phi^{\star}$ be monotonically decreasing, but it does encourage the ROA estimate to not shrink below the initial ROA estimate.

\subsubsection{Residual Loss}
\label{SEC: Residual Loss}
Another loss term that we use is the residual loss defined by

\begin{align}
    L_{r} = \sqrt{\sum_{i = 1}^{n_{c}} \left( r(s_{c, i}) \right)^2}
\end{align}

\noindent where $s_{c, i} \in \mathcal{S}$ and $n_{c} \in \N$ is again the total number of collocation training points.
The residual loss enforces agreement between the ANN's $\phi^{\star}$ estimate and the Yuan-Li PDE at the specified collocation points.

\subsubsection{Variational Loss}
\label{SEC: Variational Loss}
The most computationally intensive loss term that we use is the variational loss defined by

\begin{align}
    L_{v} = \sqrt{ \sum_{i = 1}^{n_{g}} v_{g_i}^2 }
\end{align}

\noindent where $n_{g} \in \N$ is the number of compactly supported test functions used to assess the variational loss.
While the definition of the variational loss is simple, the computational complexity comes from the evaluation of $v_{g_i}$.
In order to assess the variational loss $v_{g_i}$ associated with each test function $g_i(s)$, we make a variety of simplifications that allow us to apply Gauss-Legendre quadrature of order $n_{o} \in \N$.
The procedure that we use here is similar to that described by \cite{khodayi-mehr_varnet_2019}.
Suppose that for each collocation training point $s_{c, i} \subset \mathcal{S}$ we construct a cuboid finite element $\mathcal{S}_i \subset \mathcal{S}$ with side length $\sigma_i \in \R_{+}$ and oriented such that the element is aligned with the global coordinate frame (this is not strictly necessarily, but simplifies the proceeding analysis).
Note that these elements are only truly cubes in three dimensions -- they are line elements in one dimension, squares in two dimensions, and hyper cuboids in greater than three dimensions.
Each of these elements contains $n_{e} = n_{o}^{d}$ evenly-spaced points and the same number of standard basis functions.
We will choose our compactly supported functions $g_i(s)$ to be the standard linear basis functions of these elements, resulting in a total of $n_{g} = n_{e} n_{c}$ such functions.
To aid our upcoming analysis, we will define a secondary indexing of the support functions $g_i(s) = g_{jk}(s)$ where $i = j n_{e} + k$.
The new index $j$ indicates the element to which the basis function is associated and $k$ indicates the specific basis function associated with this element. 

Since each basis function $g_{jk}(s)$ is compactly supported over their associated element $\mathcal{S}_i$, they are by definition zero everywhere else in the domain $\mathcal{S}$.
This means that the variational loss associated with a given test function $g_{jk}(s)$ needs only to be evaluated over the associated element $\mathcal{S}_i$, rather than the complete domain $\mathcal{S}$.
So, the variational loss for the test function $g_{jk}(s)$ can be computed by

\begin{align}
    v_{g_{jk}} = \int_{\mathcal{S}_j} g_{jk}(s) r(s) ds 
\end{align}

\noindent To apply Gauss-Legendre quadrature to evaluate the above integral, we must perform a change of variables to map the integration domain to $[-1, 1]$ in each spatiotemporal dimension.
Let $\Xi \subset \R^d$ such that $\Xi = [-1, 1]^d$, by which we mean $\Xi = [-1, 1] \bigtimes \cdots \bigtimes [-1, 1]$ (i.e., the $d$ times Cartesian product of the interval $[-1, 1]$).
With this definition, $\Xi$ is a cuboid centered at the origin with side length two.
This is exactly the integration domain we need to apply Gauss-Legendre quadrature.
Having intentionally chosen our elements $\mathcal{S}_j$ to be cuboid, we know that $\forall \mathcal{S}_j \subset \mathcal{S}$, $\exists h_j: \mathcal{S}_j \to \Xi$.
In fact, $h_j$ is a linear transformation comprised of scaling and translation operations.
This means that we can represent the function $h_j$ by a transformation matrix $H_j \subset \R^{d \bigtimes d}$ given by

\begin{align}
    H_j = T_j S_j
\end{align}

\noindent where $T_j$ is the translation matrix

\begin{align}
    T_j = \mat{ I_{d} & x_{c, j} \\ 0 & 1 }
\end{align}

\noindent and $S_j$ is the scaling matrix

\begin{align}
    S_j = \mat{ \left( \frac{\sigma_j}{2} \right) I_{d} & 0 \\ 0 & 1 }
\end{align}

\noindent with $I_{d}$ being the $d$-dimensional identity matrix.
The function $h_j$ associated with the transformation matrix is $h_j(s) = H_j^{-1} s$.
So, points $s \in \mathcal{S}_j$ can be related to points $\xi \in \Xi$ by $s = H_j \xi$.
Performing this change of coordinates, the variational loss associated with the support function $g_{jk}(s)$ becomes

\begin{align}
    v_{g_{jk}} = \int_{\mathcal{S}_j} g_{jk}(s) r(s) ds = \int_{\Xi} g_{jk}(H_j \xi) r(H_j \xi) |J_j| d\xi
\end{align}

\noindent where $J_j$ is the Jacobian associated with our linear mapping $H_j$.
In this case, $J_j = \left( \frac{\sigma_j}{2} \right) I_{d}$.
So $|J| = \frac{\sigma_j d}{2}$.
Since the integration domain $\Xi$ is a valid domain for Gauss-Legendre quadrature, we can approximate the above integral by

\begin{align}
    v_{g_{jk}} \approx \left( \frac{\sigma_j d}{2} \right) \sum_{i = 1}^{n_e} w_{i} g_{jk}(H_j \xi_{e, i}) r(H_j \xi_{e, i}). 
\end{align}

\noindent where $w_i$ are the Gauss-Legendre quadrature weights and $\xi_{e, i} \in \Xi$ are the $n_e$ Gauss-Legendre quadrature points that comprise the element.
Finally, let $\bar{g}_{k}(\xi) = g_{jk}(H_j \xi)$.
With this definition, $\bar{g}_{k}(\xi)$ represents each basis function in the local element coordinates, which is a more natural way of evaluating the basis functions.
This simplification yields the final result

\begin{align}
    v_{g_{jk}} \approx \left( \frac{\sigma_j d}{2} \right) \sum_{i = 1}^{n_e} w_{i} \bar{g}_{k}(\xi_{e, i}) r(H_j \xi_{e, i}). 
\end{align}

\subsubsection{Regularization Loss}
\label{SEC: Regularization Loss}
The final loss term that we use is the regularization loss defined as

\begin{align}
    L_{reg} = L_{reg, weights} + L_{reg, biases},
\end{align}

\noindent with

\begin{align}
    L_{reg, weights} &= \sum_{i = 1}^{n_l} ||\omega_{i}||_2, \\
    L_{reg, biases} &= \sum_{i = 1}^{n_l} ||b_i||_2,
\end{align}

\noindent where $\omega_{i} \in \R^{n_{width}^2}$ are the weights of the $i$th layer reshaped into a vector, $b_i \in \R^{n_{width}}$ are the biases of the $i$th layer, and $n_l \in \N$ is the number of layers. 

\subsubsection{Total Loss}
\label{SEC: Total Loss}
With all of these loss terms defined, we can define the complete loss function as

\begin{align}
    \label{EQ: Loss Function}
    L = c_{IC} L_{IC} + c_{BC} L_{BC} + c_{Mon} L_{Mon} + c_{r} L_{r} + c_{v} L_{v} + c_{reg} L_{reg}
\end{align}

\noindent where $c_{IC}, c_{BC}, c_{Mon}, c_{r}, c_{v}, c_{reg} \in \R_{+}$ are coefficients that weigh the various loss terms.

\subsection{Training Data \& Minibatching}
\label{SEC: Training Data Minibatching}
The above loss terms require several different sets of training data, including: initial condition training data $\mathcal{S}_{IC} = \X \bigtimes \{0\} \subset \mathcal{S}$, boundary condition training data $\mathcal{S}_{BC} = \partial \X \bigtimes \T \subset \mathcal{S}$, and collocation training data $\mathcal{S}_{c} \subset \mathcal{S}$.
All of these sources of training data are subsets of our spatiotemporal domain $\mathcal{S}$.
In order to ensure that we cover the spatiotemporal domain, a subset of each of these sets of training data are generated on a fixed grid with spacing $\Delta x \in \R_{+}^{d_s}$, $\Delta t \in \R_{+}$.
The rest of the training data in each of these sets is randomly distributed throughout the spatiotemporal domain, ensuring that we are not limiting our network predictions to a fixed grid.
The resolution of the spatiotemporal grid and the number of randomly distributed training points varies depending on the specific dynamical system under investigation and will be reported separately for each of the examples for which we present results.
Evaluation data is generated in the same way as the training data, namely by sampling off of a fixed spatiotemporal grid and supplementing these fixed points with randomly distributed points.

In order to make training over a large quantity of data more feasible, we employ minibatching.
Since the IC training data set, BC training data set, and collocation training data set all have different numbers of data points, we chose to use half of the total IC and BC training data and approximately $5\%$ of the total collocation training data every minibatch.
This means that we use approximately $20$ minibatches per epoch with most trainings lasting approximately $5000$ epochs.
Each set of training data is shuffled between epochs.

\subsection{Initial Conditions}
\label{SEC: Initial Conditions}
One nuance in this method that we have yet to address is that there are actually two relevant notions of ``initial conditions'' at play.
The first notion of ``initial condition'' is that which we have focused on so far: the IC $\phi_0(x): \X \to \R$ associated with the Yuan-Li PDE that is required to generate a specific solution.
As stated in the Yuan-Li theorem, given an equilibrium point $x_e \in \X^o$, this IC must be bounded, be Lipschitz continuous, and satisfy $x_e \in \mathcal{L}_{0, \phi_0} \subset \mathcal{R}_{x_e}$.
This means that the IC $\phi_{0}$ must itself be an appropriate function whose associated ROA contains the equilibrium point that is under investigation while also containing no unstable states.
While there are many different ICs that can be used for this purpose, and there are even analytical methods one could use to generate an appropriate IC given a known dynamical system, it is usually sufficient to choose a function that yields a ``circular'' ROA of the appropriate size centered at the equilibrium point.
To this end, we use a modified sigmoid curve which allows easy placement of a basin surrounding the equilibrium point and a tall plateau beyond the desired initial ROA boundary.
This sigmoid curve is $\phi_0(x): \X \to \R$ defined by

\begin{align}
    \label{EQ: IC Function}
    \phi_0(x) = \frac{a}{1 + e^{-m \left( ||x||_2 - r \right)}} + c
\end{align}

\noindent where $a, m, r, c \in \R$ are constants that represent the amplitude, slope, radius, and offset of the sigmoid curve, respectively.
These constants are chosen to adjust the IC to the specific problem.

The second notion of ``initial condition'' is that which applies to the network's weights and biases.
For new training sessions, we set the network's weights and biases using Xaiver initialization \cite{glorot_understanding_2010}.
However, it is also possible to initialize a network with pre-trained weights and biases if such parameters are available.

Despite being ostensibly small, we shall see that this difference in initial conditions yields an important distinction between traditional numerical ROA solutions and the neural ROA solution we present here.
While both the numerical and neural methods benefit from the intelligent selection of the PDE IC $\phi_{0}(x): \X \to \R$, the ROA estimate $\mathcal{R}_{u_{x_e}}^{\star}$ for either approach is inherently constrained to only expand from the initial ROA $\mathcal{R}_{\phi_{0, x_e}}$ (i.e., $\mathcal{R}_{\phi_{0, x_e}} \subset \mathcal{R}_{u_{x_e}}^{\star}$).
Since the only means through which the numerical method can transfer information between related problems is through the PDE IC $\phi_{0}(x): \X \to \R$, this means that the numerical method can not transfer information from a problem with a larger complete ROA to a problem with a smaller complete ROA. 
On the other hand, the neural network is capable of transferring information between related problems through both the PDE IC $\phi_{0}(x): \X \to \R$ and the network's weights $W_i \in \R^{n_{width} \bigtimes n_{width}}$ and biases $b_i \in \R^{n_{width}}$.
By storing information in its weights and biases, our safety network is able to quickly generate ROA estimates for related problems, even if the complete ROA of the modified dynamical system is smaller than that of the initial dynamical system.
This is due in part to the fact that the safety network learns to estimate the Yuan-Li PDE solution $\phi(x, t): \X \bigtimes \T \to \R$, which describes the evolution of the PDE IC $\phi_{0}(x): \X \to \R$ to the final implicit ROA representation function $u_{x_e}(x): \X \to \R$, not just the final implicit representation function itself.

\subsection{Experimental Structure}
\label{SEC: Experimental Structure}
Having fully detailed our approach for training safety networks to estimate the ROAs of autonomous dynamical systems, we now layout the experiments that we have conducted to verify the efficacy of this method and demonstrate its ability to transfer information between related problems.
We sort our experiments into two categories: (1) ``methodological efficacy'' experiments and (2) ``warm start'' experiments.
Each of these experiments involve applying our safety network method to different autonomous dynamical systems that illustrate the applicability and usefulness of the method.

\subsubsection{Experiment 1: Methodological Efficacy}
\label{SEC: Experiment 1: Methodological Efficacy}
The first set of example dynamical systems that we investigate are chosen to verify that our method produces the expected ROA estimates across a variety of problems.
The examples we consider for this purpose are detailed below.

\begin{example}[Closed ROA]
    \label{EX: Closed ROA}
    Consider the dynamical system $\dot{x} = f(x)$ with flow $f: \X \subset \R^{2}$ defined as

    \begin{align}
        \dot{x}_1 &= - \sin(x_1) ( -0.1 \cos(x_1) - \cos(x_2) ), \\
        \dot{x}_2 &= - \sin(x_2) ( \cos(x_1) - 0.1 \cos(x_2) ).
    \end{align}

    \noindent We wish to compute the complete ROA $\mathcal{R}_{x_e}$ associated with the equilibrium point $x_e = \mat{ \frac{\pi}{2} & \frac{\pi}{2} }^T$.
    This example is of particular interest because it features infinitely many repeating equilibria, each with their own ROAs arranged in a pattern across the state space.
    By generating a solution to this problem, we show that our safety network method is able to differentiate between the ROAs of nearby equilibrium points.
\end{example}

\begin{example}[Simple Pendulum]
    \label{EX: Simple Pendulum}
    Consider the dynamical system $\dot{x} = f(x)$ with flow $f: \X \subset \R^{2}$ defined as

    \begin{align}
        \dot{x}_1 &= x_2, \\
        \dot{x}_2 &= - \frac{g}{L} \sin x_1 - \frac{c}{m L^2} x_2,
    \end{align}

    \noindent where $m = 0.097 \, \units{kg}$, $c = 0.0024 \, \units{\frac{Ns}{m}}$, $L = 0.2 \, \units{m}$, and $g = -9.81 \, \units{\frac{m}{s^2}}$.
    We wish to compute the complete ROA $\mathcal{R}_{x_e}$ associated with the equilibrium point $x_e = \mat{ 0  & 0 }^T$.
    This dynamical system describes the behavior of an unforced simple pendulum with no spring element, generating a flow field with infinitely many alternating stable and unstable equilibria.
    By generating a solution to this problem, we show that our safety network method is able to capture the unbounded ROA of a stable equilibrium point while excluding nearby unstable equilibria.
\end{example}

\begin{example}[Cart Pendulum]
    \label{EX: Cart Pendulum}
    Consider the dynamical system $\dot{x} = f(x)$ with flow $f: \X \subset \R^{4} \to \R^4$ defined as

    \begin{align}
        \dot{x}_1 &= x_2 \\
        \dot{x}_2 &= \frac{6 c_2 x_4 \cos \left( x_3 \right) + 6 k_2 x_3 \cos \left( x_3 \right) - 4 L c_1 x_2 - 4 L k_1 x_1 + 2 L^2 m_2 x_4^2 \sin \left( x_3 \right) + 3 m_2 g L \cos \left( x_3 \right) \sin \left( x_3 \right)}{L \left( 4 ( m_1 + m_2 ) - 3 m_2 \cos^2 \left( x_3 \right) \right)} \\
        \dot{x}_3 &= x_4 \\
        \dot{x}_4 &= - \frac{ -6 m_2 L \cos ( x_3 ) \left( k_1 x_1 + c_1 x_2 \right) + 12 ( m_1 + m_2 ) \left( k_2 x_3 + c_2 x_4 \right) + \frac{3}{2} m_2^2 L^2 \sin(2 x_3) x_4^2 + 6 m_2 ( m_1 + m_2 ) g L \sin (x_3) }{ m_2 L^2 \left( 4 ( m_1 + m_2 ) - 3 m_2 \cos^2 \left( x_3 \right) \right)}
    \end{align}

    \noindent where $m_1 = 0.257 \, \units{kg}$, $m_2 = 0.127 \, \units{kg}$, $c_1 = 0.0024 \, \units{\frac{Ns}{m}}$, $c_2 = 0.0024 \, \units{\frac{Ns}{m}}$, $k_1 = 0.1 \, \units{\frac{N}{m}}$, $k_2 = 0 \, \units{\frac{N}{m}}$, $L = 0.3365 \, \units{m}$, and $g = -9.81 \, \units{\frac{m}{s^2}}$.
    We wish to compute the complete ROA $\mathcal{R}_{x_e}$ associated with the equilibrium point $x_e = \mat{ 0  & 0 }^T$.
    This dynamical system describes the behavior of an unforced cart pendulum with a spring element on the cart, generating a flow field similar to that of Example \ref{EX: Simple Pendulum} when projected on the pendulum states.
    Correctly estimating the ROA of this system demonstrates the ability of our method to estimate the ROA of higher dimensional non-linear autonomous dynamical systems.
\end{example}

\subsubsection{Experiment 2: Warm Start}
\label{SEC: Experiment 2: Warm Start}
The second set of experiments that we conduct involve verifying that our safety network method is able to quickly learn to estimate the ROAs of related problems by transferring information from previous examples.
These experiments emphasize the advantage of storing information in weights and biases, which allows our safety network to quickly learn related solutions without having to modify the underlying PDE IC.
As noted earlier, this is particularly important for our application of ROA estimation, because transferring information via PDE ICs is only viable when the complete ROA of the new system under investigation contains the complete ROA of a previous system.
To this end, we pre-train our safety network to estimate the ROA of Example \ref{EX: Simple Pendulum}, which we call variation ``2a'', and then conduct further trainings on two different variations ``2b,'' and ``2c'' of the same simple pendulum problem.
These simple pendulum variations have the same underlying equations of motion, but use different pendulum lengths as documented in Table \ref{TABLE: Simple Pendulum Variations}.
As we shall see in the results section, the first of these parameter modifications does not significantly impact the shape of the flow field, while the second has a much more significant impact.

\begin{table}[htbp]
    \caption{Warm start simple pendulum variations.}
    \label{TABLE: Simple Pendulum Variations}
    \centering
    \begin{tabular}{ |c|c|c|c|c| }
        \hline
        ID & Example & Mass & Damping & Length \\
        \hline
        [\#] & [-] & [kg] & [Ns/m] & [m] \\
        \hline
        1 & 2a & 0.127 & 0.0024 & 0.2 \\
        2 & 2b & 0.127 & 0.0024 & 0.3 \\
        3 & 2c & 0.127 & 0.0024 & 0.4 \\
        \hline
    \end{tabular}
\end{table}

\section{Results}
\label{SEC: Results}
Having laid out the example dynamical systems of interest to us, we now present a variety of results that demonstrate the success of our safety network method at estimating the associated complete ROAs.
For each example, we provide: (1) the analytical PDE IC that we use to guide the training; (2) the training evaluation loss vs epoch number; (3) the safety network estimate of the initial and final implicit ROA surfaces; and (4) the flow field of the dynamical system with the initial and final safety network ROA estimates, as well as a high fidelity numerical ROA estimate.
In each case, the training evaluation loss indicates that our safety network has converged to a solution, while the ROA boundary plots confirm that the safety network estimate reasonably agrees with the numerical solution.

\subsection{Experiment 1: Methodological Efficacy}
\label{SEC: Experiment 1: Methodological Efficacy Results}
First, we show results indicating that our safety network approach succeeds at estimating the complete ROAs of various autonomous dynamical systems using the examples outlined in section \ref{SEC: Experimental Structure}.

\subsubsection{Example \ref{EX: Closed ROA}: Closed ROA}
\label{SEC: Closed ROA Example Results}
While this autonomous dynamical system has infinitely many stable equilibria patterned throughout the state space, we focus our attention on estimating the ROA for the equilibrium point $x_e = \mat{ \frac{\pi}{2} & \frac{\pi}{2} }^T$ located at the origin.
For this problem, we choose a spatial domain of $\X = [ -1, 4 ] \bigtimes [ -1, 4 ]$ and a temporal domain of $\T = [ 0, 30 ]$, with a spatial discretization of $\Delta x = 0.6319$ and a temporal discretization of $\Delta t = 0.5263$, respectively.
Let $\mathcal{S} = \X \bigtimes \T$ be the complete spatiotemporal domain.
We randomly sample an additional $10000$ collocation points from $\mathcal{S}$.
Note that an appropriate selection of the spatiotemporal domain requires some apriori knowledge of the system being studied.
Since the network only learns from samples taken from the spatiotemporal domain $\mathcal{S}$, it is clearly important that the spatial domain $\X$ contains the complete ROA $\mathcal{R}_{x_e}$ (i.e., it is necessary that $\mathcal{R}_{x_e} \subset \X$).
At the same time, however, choosing an unnecessarily large spatial domain $\X$ will result in a smaller number of training samples taken from the relevant region of the state space near the complete ROA, resulting in either an inaccurate answer or an unnecessarily long training duration.
Similarly, the solution to the Yuan-Li PDE is only guaranteed to converge to the necessary Lyapunov function as $t \to \infty$.
As such, it is necessary to choose the time domain $\T \subset \R_{+}$ so that the Yuan-Li PDE has enough time to converge to the correct Lyapunov function while not making the training continue for an implausibly long span of time.
Fortunately, the Yuan-Li PDE converges to the desired implicit ROA representation relatively quickly, though this is problem dependent.

\begin{figure}[htbp]
	\centering
  	\includegraphics[width=4.5in]{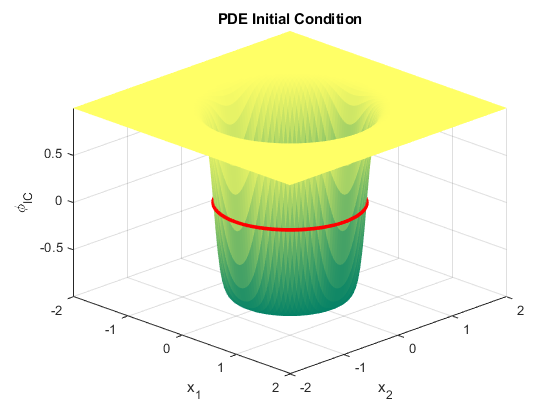}
  	\caption{Yuan-Li PDE initial condition surface.}
    \label{FIG: PDE IC}
\end{figure}

To initialize our safety network, we use the PDE IC shown in Fig. \ref{FIG: PDE IC}, simply shifted to be centered at the equilibrium point $x_e = \mat{ \frac{\pi}{2} & \frac{\pi}{2} }^T$.
This is an example of the surface described by the aforementioned IC function in Eq. \ref{EQ: IC Function}.
Our other examples all also use this IC without needing to shift it, since their equilibrium points are located at or near the origin.
This IC is chosen such that the associated ROA contains the equilibrium point of interest but not any unstable regions of the state space.

\begin{figure}[htbp]
	\centering
  	\includegraphics[width=6in]{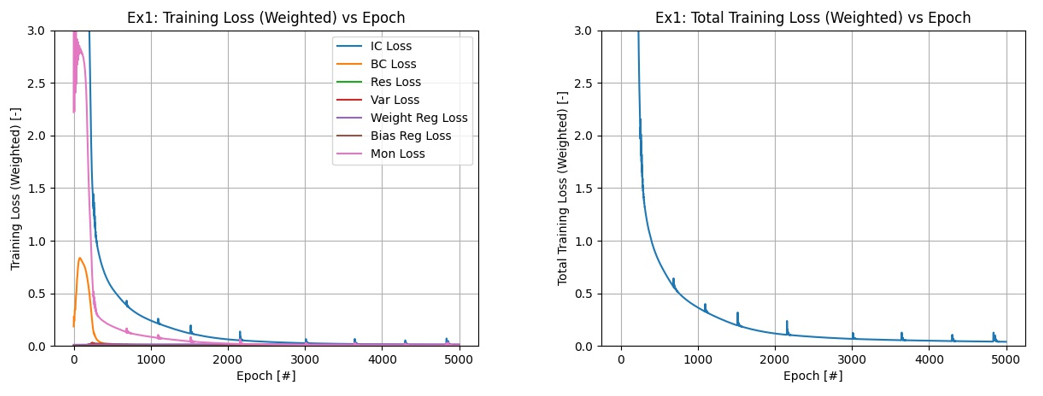}
  	\caption{(a) Ex1 Training Loss Components. (b) Ex1 Total Training Loss.}
    \label{FIG: Ex1 Training Losses}
\end{figure}

Since this example yields a closed ROA, we choose to enforce the boundary conditions that accompany the given PDE IC.
For examples where we expect to have open boundary conditions, we instead use free boundary conditions to all the ROA estimate to extend to the edge of the domain.
With the training domains defined and the PDE ICs and BCs established, we can now train this network by minimizing our loss function Eq. \ref{EQ: Loss Function}. 
The evaluation losses per epoch are shown in Fig. \ref{FIG: Ex1 Training Losses}.
These two plots demonstrate that our network has successfully been optimized to minimize the various terms in our loss function.

\begin{figure}[htbp]
	\centering
  	\includegraphics[width=6in]{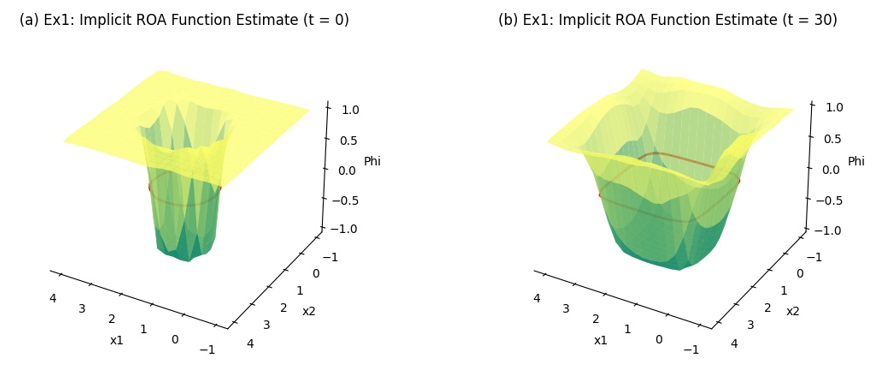}
  	\caption{(a) Ex1 Implicit ROA Function Estimate at $t = 0$. (b) Ex1 Implicit ROA Function Estimate at $t = 30$.}
    \label{FIG: Ex1 ROA Function Estimates}
\end{figure}

Before training, the safety network's ROA predictions are essentially random surfaces.
After training, however, we can view the quality of our network's ROA estimation by examining the initial and final implicit ROA functions generated by the network and comparing the complete ROA estimates associated with each of these functions to numerical results.
The initial and final ROA estimate functions generated by our trained safety network can be seen in Fig. \ref{FIG: Ex1 ROA Function Estimates}.
While it can be difficult to read the ROA boundary from these 3D surface plots, we include them to show how the network evolves the implicit ROA function over time in accordance with the Yuan-Li PDE.

\begin{figure}[htbp]
	\centering
  	\includegraphics[width=6in]{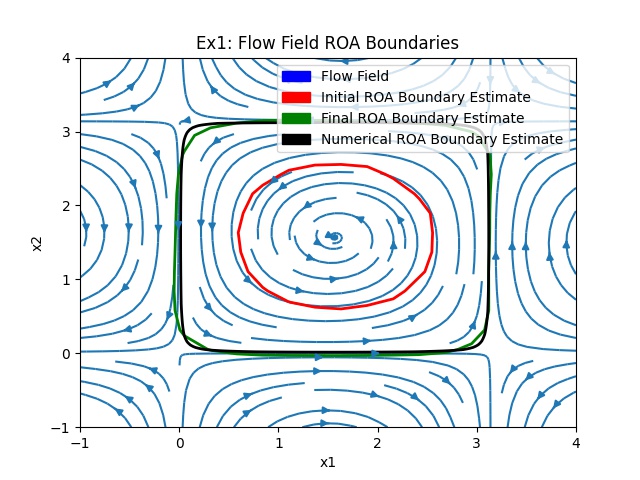}
  	\caption{Ex1 Flow Field with ROA Boundaries.}
    \label{FIG: Ex1 Flow Field}
\end{figure}

By plotting the zero level set of the PDE solution produced by our safety network at different time steps as contours over the dynamical system's flow field, we can illustrate the evolution of our ROA prediction over time.
For simple flow fields, the complete ROA estimate can sometimes be verified by visual inspection of the flow field, however this becomes less plausible for high dimensional problems, so we provide numerical solutions for the purpose of comparison.
Fig. \ref{FIG: Ex1 Flow Field} shows how our neural ROA estimates evolve over time.
Note that our safety network ROA estimate agrees well with the numerical estimate of the ROA.

\subsubsection{Example \ref{EX: Simple Pendulum}: Simple Pendulum}
\label{SEC: Simple Pendulum Example Results}
\begin{figure}[htbp]
	\centering
  	\includegraphics[width=6in]{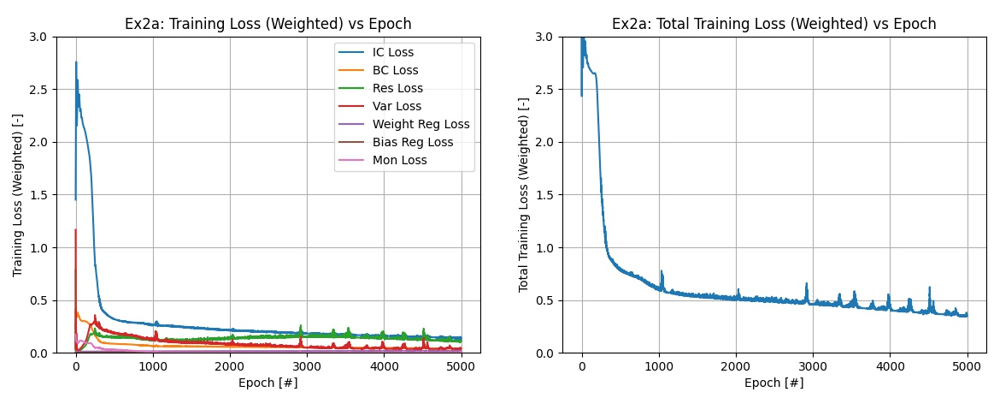}
  	\caption{(a) Ex2a Simple Pendulum Training Loss Components. (b) Ex2a Simple Pendulum Total Training Loss.}
    \label{FIG: Ex5a Training Losses}
\end{figure}

The unforced simple pendulum with no spring element has a flow field characterized by infinitely many equilibria arranged along the x-axis of the state space.
These equilibria alternate between being stable and unstable, with the stable equilibria possessing unbounded ROAs that cover nearly the entire state space.
For this problem, we have aligned the coordinate system of our dynamical system such that the system has a stable equilibrium point at $x_e = \mat{ 0 & 0 }^T$.
Due to the conspicuous pattern in the flow field, it is easy to pick a spatial domain that includes all of the relevant portions of the state space.
In this case, we choose a spatial domain of $\X = [-2 \pi, 2 \pi] \bigtimes [ -4 \pi, 4 \pi ]$ and a temporal domain of $\T = [0, 10]$, with the same spatiotemporal discretizations as before (i.e., $\Delta x = 0.6319$ and $\Delta t = 0.5263$).
It should be noted that, since our network learns only from data within the spatial domain $\X$, our network cannot capture stability behavior that occurs outside of this domain.
For systems with unbounded ROAs such as this one, this means that the safety network's ROA prediction extends to the boundary of the spatial domain $\partial \X$ without forming a closed shape.
As before, we randomly select an additional $10000$ collocation points from the spatiotemporal domain $\mathcal{S} = \X \bigtimes \T$.
For this example we use the exact PDE initial condition shown in Fig. \ref{FIG: PDE IC} (without shifting it) while we opt to use a free boundary condition due to the open nature of the ROA that we expect.
The evaluation losses per epoch are shown in Fig. \ref{FIG: Ex5a Training Losses}.
These evaluation losses once again indicate that our safety network has converged to a solution that minimizes our total loss function.

\begin{figure}[htbp]
	\centering
  	\includegraphics[width=6in]{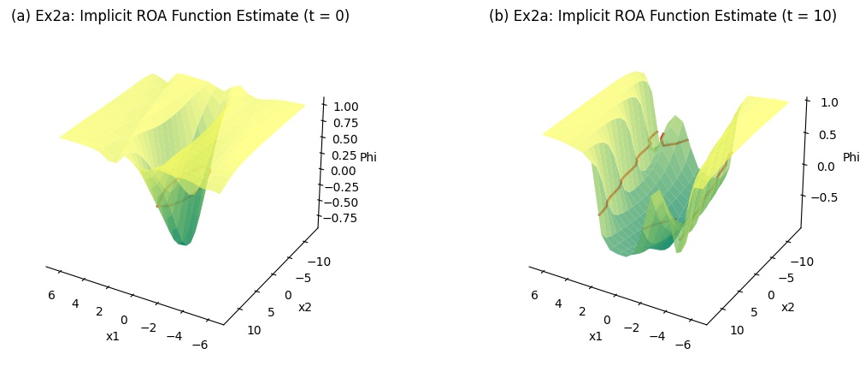}
  	\caption{(a) Ex2a Simple Pendulum ROA Function Estimate at $t = 0$. (b) Ex2a Simple Pendulum ROA Function Estimate at $t = 10$.}
    \label{FIG: Ex5a Lyapunov Function Estimates}
\end{figure}

To show how the implicit ROA function evolves over time, we once again provide the initial and final snapshots of our network estimate (see Fig. \ref{FIG: Ex5a Lyapunov Function Estimates}). 
These plots reveal more detail about the safety network estimates, but are difficult to use as tools to evaluate the quality of the associated ROA boundary.

\begin{figure}[htbp]
	\centering
  	\includegraphics[width=6in]{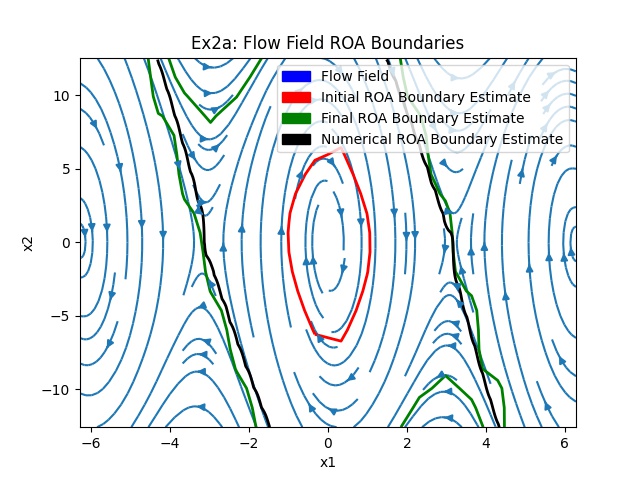}
  	\caption{Ex2a Simple Pendulum Flow Field With ROA Estimations.}
    \label{FIG: Ex5a ROA Boundary Prediction}
\end{figure}

To assess the quality of our safety network estimate, we show the provided the safety network ROA boundaries overlaid onto the flow field of the dynamical system in Fig. \ref{FIG: Ex5a ROA Boundary Prediction}.

As before, we see that our safety network estimate reasonably approximates the ROA boundary predicted by the numerical solution.
The fluctuations in the prediction along the boundary are due to the sharp discontinuity at the boundary compared to the resolution of our evaluation data.

\subsubsection{Example \ref{EX: Cart Pendulum}: Cart Pendulum}
\label{SEC: Cart Pendulum Example Results}
\begin{figure}[htbp]
	\centering
  	\includegraphics[width=6in]{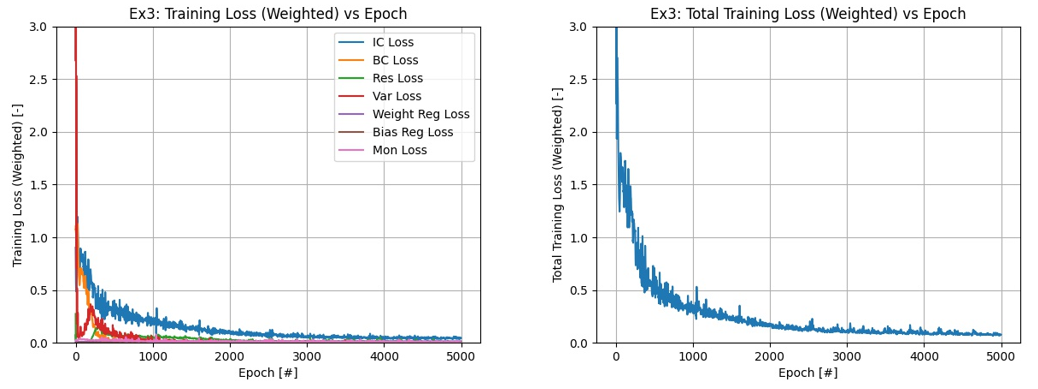}
  	\caption{(a) Ex3 Cart Pendulum Training Loss Components. (b) Ex3 Cart Pendulum Total Training Loss.}
    \label{FIG: Ex11 Training Losses}
\end{figure}

Like the simple pendulum, the cart pendulum has a flow field characterized by an alternating pattern of stable and unstable equilibria.
Unlike the simple pendulum, however, the cart pendulum is a higher dimensional problem with four state variables, making it difficult to illustrate the complete flow field and the associated ROAs in a single image.
Despite the higher dimensionality of this problem, it is still possible to use relevant background knowledge to generate appropriate spatiotemporal training domains and to estimate the ROA shape we expect.
The first spatial state $x_1 \in \R$ is the cart position, which in the standard cart pendulum setup will be constrained by the length of the rail on which the cart moves.
Similarly, the second spatial state $x_2 \R$, the cart velocity, will also be limited by the physical hardware, such as the maximum velocity of the motors that actuate the cart.
Finally, the third and fourth state variables $x_3, x_4 \in \R$, representing the pendulum angle and angular velocity, respectively, may each be bounded in the same way as for the simple pendulum.
Since we do not have a particular hardware setup to analyze, we are free to choose reasonable domains that simplify the problem.
In this case, we select a spatial domain of $\X = [ -2 \pi, 2 \pi ] \bigtimes [ -4 \pi, 4 \pi ] \bigtimes [ -2 \pi, 2 \pi ] \bigtimes [ -4 \pi, 4 \pi ]$ and a temporal domain of $\T = [ 0, 10 ]$, with the same spatiotemporal discretizations as before (i.e., $\Delta x = 0.6319$ and $\Delta t = 0.5263$). 
Again, we randomly select an additional $10000$ collocation points from the spatiotemporal domain $\mathcal{S} = \X \bigtimes \T$ and use the same PDE ICs and BCs as in Ex. \ref{EX: Closed ROA}.
The evaluation losses per epoch are shown in Fig. \ref{FIG: Ex11 Training Losses}.
In this case we also see good convergence of our the terms in our loss function,

\begin{figure}[htbp]
	\centering
  	\includegraphics[width=6in]{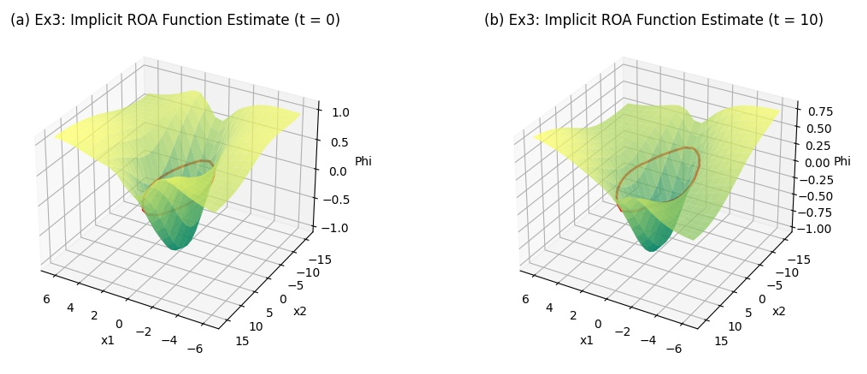}
  	\caption{(a) Ex3 Cart Pendulum ROA Function Estimate at $t = 0$. (b) Ex3 Cart Pendulum ROA Function Estimate at $t = 10$.}
    \label{FIG: Ex11 Lyapunov Function Estimates}
\end{figure}

\begin{figure}[htbp]
	\centering
  	\includegraphics[width=6in]{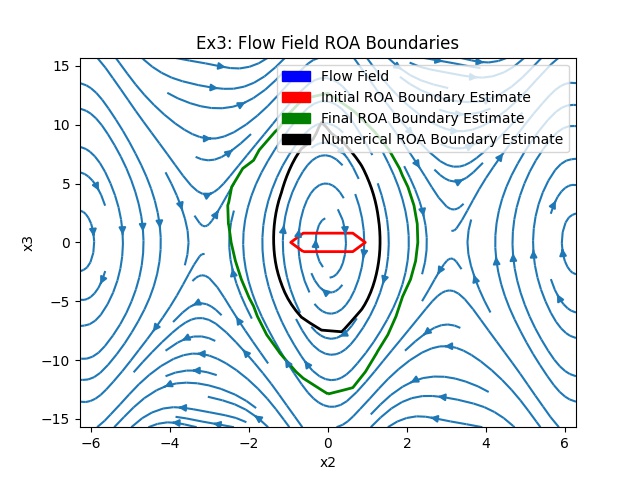}
  	\caption{Ex3 Cart Pendulum Flow Field with ROA Boundaries.}
    \label{FIG: Ex11 Flow Field}
\end{figure}

To evaluate the quality of our ROA estimate, we once again provide both the initial and final implicit ROA function estimates (see Fig. \ref{FIG: Ex11 Lyapunov Function Estimates}) as well as the associated ROA boundaries overlaid onto the flow field of the dynamical system (see Fig. \ref{FIG: Ex11 Flow Field}).
Due to the higher dimensionality of the state space of this problem, we represent the flow field, implicit ROA functions, and associated ROAs as projections on pairs of state variables.
Since we have included a spring element on the cart, projecting the dynamics onto the cart states simply yields a global sink, which does not reveal much information about the stability of the system (since these states are globally stable).
Instead, we choose to project the dynamics onto the pendulum states, as shown in Figs. \ref{FIG: Ex11 Lyapunov Function Estimates} and \ref{FIG: Ex11 Flow Field}, since these are the most informative states.
This projection choice also reveals the similarity of the cart pendulum to the simple pendulum in the previous example.
As before, the safety network estimates of the ROA boundary agree fairly well with the numerical estimate.

\subsection{Experiment 2: Warm Start}
\label{SEC: Experiment 2: Warm Start Results}
The second set of experimental results that we show demonstrate that our safety network is able to transfer information between related problems, allowing the network to quickly generate ROA estimates to dynamical systems that are slight modifications of one another.
We do this by pre-training our safety network on simple pendulum variation ``a'' (whose results are presented in Section \ref{SEC: Simple Pendulum Example Results}) and then conducting additional trainings on simple pendulum variations ``b'' and ``c'' (whose parameters are enumerated in Table \ref{TABLE: Simple Pendulum Variations}).
Since all of these simple pendulum variations share the same underlying dynamical system, we select the same spatial and temporal domains for analysis, namely  $\X = [-2 \pi, 2 \pi] \bigtimes [ -4 \pi, 4 \pi ]$ and $\T = [0, 10]$, respectively, with $\Delta x = 0.6319$ and $\Delta t = 0.5263$.
Similarly, we use the same PDE IC shown in Fig. \ref{FIG: PDE IC}.
For each simple pendulum variation, we now present training results with and without pre-training on variation ``a.''

\subsubsection{Example \ref{EX: Simple Pendulum}b}
\label{SEC: Simple Pendulum b Results}
\begin{figure}[htbp]
	\centering
  	\includegraphics[width=6in]{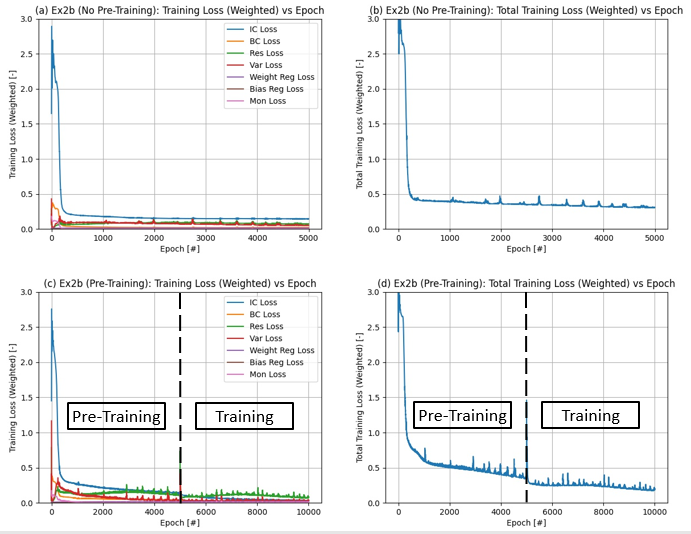}
  	\caption{(a) Ex2b Simple Pendulum Training Loss Components Without Pre-Training. (b) Ex2b Simple Pendulum Total Training Loss Without Pre-Training. (c) Ex2b Simple Pendulum Training Loss Components With Pre-Training. (d) Ex2b Simple Pendulum Total Training Loss With Pre-Training.}
    \label{FIG: Ex5b Training Losses}
\end{figure}

The evaluation losses per epoch for simple pendulum variation ``b'' with and without pre-training on variation ``a'' are shown in Fig. \ref{FIG: Ex5b Training Losses}.
This plot indicates that, while both the pre-trained and untrained models converge to a result, the pre-trained model converges to a result more quickly than the untrained model.
This is particularly true for this example, since increasing the length of the pendulum by $50\%$ did not have a significant impact on the shape of the flow field.

\begin{figure}[htbp]
	\centering
  	\includegraphics[width=6in]{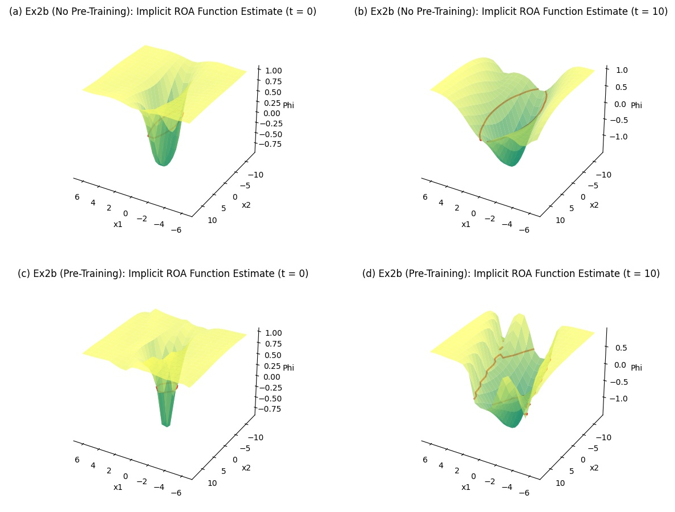}
  	\caption{(a) Ex2b Simple Pendulum Initial ROA Estimate Function Without Pre-Training. (b) Ex2b Simple Pendulum Final ROA Estimate Function Without Pre-Training. (c) Ex2b Simple Pendulum Initial ROA Estimate Function With Pre-Training. (d) Ex2b Simple Pendulum Final ROA Estimate Function With Pre-Training.}
    \label{FIG: Ex5b Lyapunov Function Estimates}
\end{figure}

\begin{figure}[htbp]
	\centering
  	\includegraphics[width=6in]{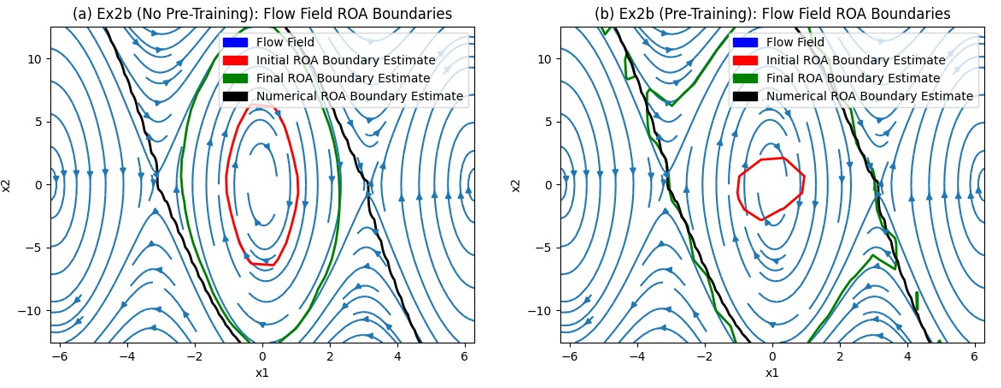}
  	\caption{(a) Ex2b Simple Pendulum Flow Field With ROA Estimations Without Pre-Training. (b) Ex2b Simple Pendulum Flow Field With ROA Estimations With Pre-Training.}
    \label{FIG: Ex5b ROA Boundary Prediction}
\end{figure}

Fig. \ref{FIG: Ex5b Lyapunov Function Estimates} compares the initial and final implicit ROA estimation functions with and without pre-training, while Fig. \ref{FIG: Ex5b ROA Boundary Prediction} compares the associated ROAS.
An important observation from the ROA boundary plots is that the network without pre-training predicts a smaller stable region than does the network that was pre-trained.
This is due to the fact that the additional states captured by the pre-trained model all have flows that exit our domain of analysis.
When this happens, it is difficult for the network to predict whether the states re-enter the domain of interest far into the future, since the network is restricted to information contained only within the given domain.
The pre-trained network is able to correctly classify these additional states as stable since these states were also stable in the original pendulum variation Ex. \ref{EX: Simple Pendulum}a.

Note that while the final results of training with and without pre-training are similar, the pre-training example converges more quickly to this final solution.
These results indicate that the safety network is able to transfer information from the pre-trained simple pendulum variation ``a'' to a new pendulum variation, in this case variation ``b.''
As we shall see, this same result holds true for more significant variations of the simple pendulum model, such as in Ex. \ref{EX: Simple Pendulum}c.

\subsubsection{Example \ref{EX: Simple Pendulum}c}
\label{SEC: Simple Pendulum c Results}
\begin{figure}[htbp]
	\centering
  	\includegraphics[width=6in]{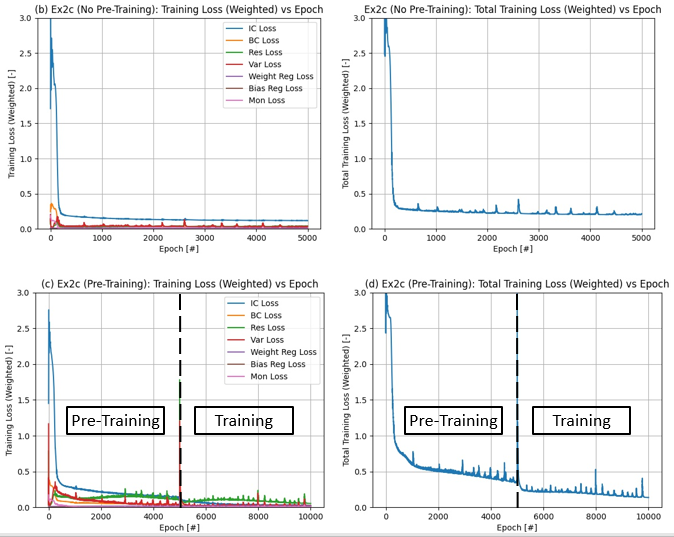}
  	\caption{(a) Ex2c Simple Pendulum Training Loss Components Without Pre-Training. (b) Ex2c Simple Pendulum Total Training Loss Without Pre-Training. (c) Ex2c Simple Pendulum Training Loss Components With Pre-Training. (d) Ex2c Simple Pendulum Total Training Loss With Pre-Training.}
    \label{FIG: Ex5c Training Losses}
\end{figure}

For simple pendulum variation ``c'' we present the same type of convergence information as the previous variation.
Fig. \ref{FIG: Ex5c Training Losses} shows the evaluation losses per epoch for simple pendulum variation ``c'' with and without pre-training on variation ``a.''

\begin{figure}[htbp]
	\centering
  	\includegraphics[width=6in]{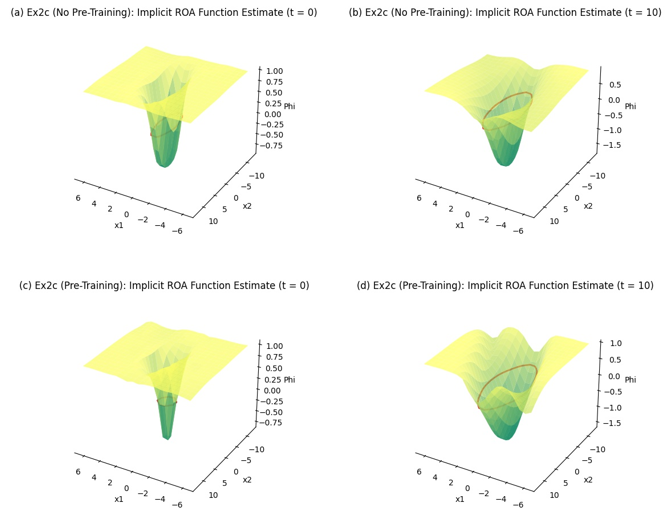}
  	\caption{(a) Ex2c Simple Pendulum Initial ROA Estimation Function Estimate. (b) Ex2c Simple Pendulum Final ROA Estimation Function. (c) Ex2c Simple Pendulum Initial ROA Estimation Function With Pre-Training. (d) Ex2c Simple Pendulum Final ROA Estimation Function With Pre-Training.}
    \label{FIG: Ex5c Lyapunov Function Estimates}
\end{figure}

\begin{figure}[htbp]
	\centering
  	\includegraphics[width=6in]{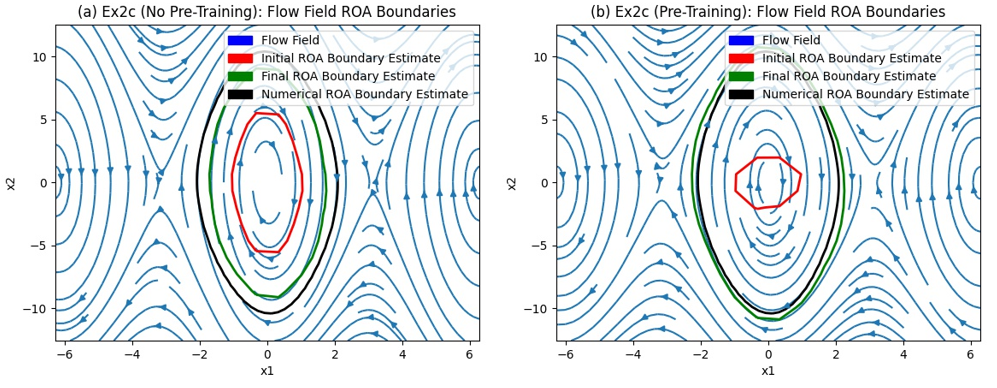}
  	\caption{(a) Ex2c Simple Pendulum Flow Field With ROA Estimations Without Pre-Training. (b) Ex2c Simple Pendulum Flow Field With ROA Estimations With Pre-Training.}
    \label{FIG: Ex5c ROA Boundary Prediction}
\end{figure}

As before, the evaluation losses of the pre-trained network converge more quickly than do the untrained network evaluation losses, and achieve an overall lower loss value, once again indicating that relevant information from variation ``a'' has been applied to a related system.
Along the same lines, we show the initial and final Lyapunov function estimates in Fig. \ref{FIG: Ex5c Lyapunov Function Estimates} and the associated ROAs in Fig. \ref{FIG: Ex5c ROA Boundary Prediction}.
In this example, doubling the pendulum length has had a much more significant impact on the shape of the flow field, decreasing the size of the complete ROA.
The pre-trained network achieve a slightly larger ROA than the network without pre-training.

\section{Discussion}
\label{SEC: Discussion}
The central goal of this work is to create a neural framework for ROA estimation that can be used to assess the stability of controlled dynamical systems, with the future application of evaluating the exploratory actions of online reinforcement learners to prevent unsafe outcomes.
While the application of our method to reinforcement based motor learning problems is left for future work, we argue that the safety network approach presented here establishes a strong foundation for PDE based neural ROA estimation with numerous possible applications and future extensions.
By combining existing Lyapunov stability theory such as the Yuan-Li theorem with neural solutions to PDEs, we are able to train networks with relatively simple architecture to estimate the ROAs of autonomous dynamical systems.
The results of our ``methodological efficacy'' experiments indicate that our safety network method is able to estimate the ROAs of a diverse set of dynamical systems, including systems with both closed and open ROAs, as well as systems of moderate dimension (e.g., four state variables and a single temporal variable in Ex. \ref{EX: Cart Pendulum}).
At the same time, our ``warm start'' experiments emphasize one of the key advantages of our neural ROA solution over traditional numerical ROA solutions; namely, that by storing information in its weights and biases, our safety network is able to transfer information between related dynamical systems without having to alter the PDE IC.
Even with relatively significant modifications of the simple pendulum parameters (e.g., doubling the pendulum length) that resulted in large changes to the shape of the system's ROA, our safety network was able to quickly learn to estimate the ROA of the new dynamical system when it was pre-trained on the initial system.
This advantage is of important for assessing the stability of dynamical systems controlled by reinforcement learners, since such control schemes slowly modify themselves over time as new experiences are accumulated, requiring the ROA estimation technique to also be able to adapt. 
Overall, the results that we have presented here validate our safety network method as a useful tool for estimating the ROA of autonomous dynamical systems.

Despite the success of this method on valid applications, it is important to point out several assumptions and limitations of this method that narrow the scope of its applicability.

\subsection{Autonomous Dynamical Systems}
Since our safety network framework is based on a solution to the Yuan-Li PDE and the Yuan-Li theorem is valid only for autonomous dynamical systems, this means that our safety network is likewise incapable of assessing non-autonomous dynamical systems, which may be of interest to certain scientific and engineering applications.
For our primary application of motor learning, this means that a fixed network is unable to assess the stability of a control system whose parameters are changing over time or whose open loop system has parameters that change over time.
This would, of course, become possible if the safety network's parameters were allowed to update as the underlying controller improved, which might be possible in an online learning scenario.

\subsection{Discrete Equilibria}
Along the same lines, our safety network method is only applicable to dynamical systems that have discrete equilibria that can be analyzed.
In each of the example dynamical systems that we analyzed here, such as the cart pendulum problem with a spring element on the cart (see Ex. \ref{EX: Cart Pendulum}), there are infinitely many discrete equilibria that are dispersed throughout the state space.
If we instead analyze the cart pendulum with no spring element, there would be infinitely many equilibria arranged arbitrarily closely to one another along a stable line through the state space.
This occurs because the cart position is a continuous variable and, without the spring element, any cart position is stable.
It therefore becomes impossible to generate a PDE IC that contains only the equilibrium point of interest and not the arbitrarily many nearby stable equilibria, causing the ROA estimate to grow well beyond what would be expected for the single equilibrium point of interest.
This is not so much an issue with the safety network methodology itself as it is one with ill-posed nature of the problem.

\subsection{Limited Domain of Analysis}
One final limitation to the applicability of our safety network method is the limitation of valid predictions to within the specified spatial domain $\X$.
Not only does this mean that it is important to make an intelligent selection for the spatial domain so that the relevant ROA is encompassed, but it also means that some states whose flow leaves the domain $\X$ may sometimes be falsely labeled unstable, even if these flows re-enter the domain at some future time.
This occurs because the safety network is limited only to information present within the specified domain, and while the PDE does provide information about the evolution of solutions over time, the network is unable to extend these predictions arbitrarily far forward in time.
As a result it is difficult for the network to make accurate predictions about the distant future trajectories of flows that leave the domain of interest.

\section{Future Work}
\label{SEC: Future Work}
As we have already mentioned, it is our intention to apply our safety network method to assess reinforcement learning agents for motor control applications such as legged locomotion.
However, the extensions of this work go well beyond this specific application.
In particular, the rapid development within the field of scientific machine learning and neural solutions to PDEs offers several different interesting extensions of this work.
As others in the field have begun to address, the basic PINN framework can be extended to better solve problems involving stiff PDEs \cite{wang_understanding_2020}, PDEs with significant discontinuities \cite{jagtap_conservative_2020,karniadakis_extended_2020}, and PDEs of high dimension, all of which are highly relevant to the problem of ROA estimation that our safety network attempts to address.
Efforts to extend PINNs to solve parameterized PDEs \cite{dal_santo_data_2020} could also offer a way for our safety network framework to assess the ROAs of whole families of dynamical systems (e.g., all pendulum problems) or families of controllers (e.g., all PID controllers for a single system).
Such extensions of the safety network method would yield a significant improvement over repeated numerical ROA analysis for different open loop or controller parameters.

\section{Conclusion}
\label{SEC: Conclusion}
As the demands and applications robotics extends to ever more complex motor tasks in unstructured environments, neural control solutions that apply a reinforcement learning framework to motor learning tasks have become a subject of greater interest.
A necessary component of the development of reinforcement learning methods for motor control is the ability to assess the stability of these techniques and to constrain these agents to take exclusively ``safe'' actions while continuing to learn in hardware. 
While the problem of safety in reinforcement learning for complex motor tasks is not a challenge that can be solved in a single paper, the safety network framework that we present in this paper offers a foundation for neural ROA estimation of autonomous dynamical systems.
In this work, we have demonstrated that a relatively simple ANN architecture is able to estimate the ROA of autonomous dynamical systems by learning to represent the solution to the Yuan-Li PDE, whose solution converges to a special function that yields the complete ROA of interest.
Although this method works across a variety of different autonomous dynamical systems and quickly learns to represent the ROAs of related dynamical systems, the safety network approach is just the beginning of our investigation into neural ROA estimation.
We intend to continue studying these topics by pursuing several of the aforementioned areas of future work, with an ultimate aim of integrating our safety network with a reinforcement learning agent to ensure safety during training.

\section{Funding}
\label{SEC: Funding}
This work was performed at the U.S. Naval Research Laboratory under the Base Program's Safe Lifelong Motor Learning (SLLML) work unit, WU1R36.

\section{Supplemental Data}
\label{SEC: Supplemental Data}
It is the intend of the authors to eventually make the associated code available after it has been formally published.

\section{Conflict of Interest Statement}
\label{SEC: Conflict of Interest Statement}
The authors declare that the research was conducted in the absence of any commercial or financial relationships that could be construed as a potential conflict of interest.

\bibliographystyle{unsrt}
\bibliography{main} 

\newpage

\appendix

\section{Appendix}
\label{SEC: Appendix}

\subsection{Table of Variables}
\label{SEC: Table of Variables}

\begin{table}[htbp]
    \caption{List of variable symbols and descriptions.}
    \label{TABLE: Variable Definitions}
    \centering
    \begin{tabular}{ |c|c|c| }
        \hline
        ID & Variable & Description \\
        \hline
        1 & $x$ & Spatial Variable \\ 
        \hline
        2 & $x_e$ & Spatial Equilibrium Point \\ 
        \hline
        3 & $t$ & Temporal Variable \\ 
        \hline
        4 & $s$ & Spatiotemporal Variable \\ 
        \hline
        5 & $\xi$ & Spatiotemporal Variable of the Set $\Xi$ \\ 
        \hline
        6 & $\sigma_i$ & Scaling Matrix Factors  \\ 
        \hline
        7 & $\X$ & Spatial Domain \\ 
        \hline
        8 & $\partial \X$ & Spatial Domain Boundary \\ 
        \hline
        9 & $\X^o$ & Spatial Domain Interior \\
        \hline
        10 & $\T$ & Temporal Domain \\ 
        \hline
        11 & $\mathcal{S}$ & Spatiotemporal Domain \\ 
        \hline
        12 & $\mathcal{S}_{IC}$ & Initial Condition Spatiotemporal Domain \\
        \hline
        13 & $\mathcal{S}_{BC}$ & Boundary Condition Spatiotemporal Domain \\
        \hline
        14 & $\mathcal{S}_{c}$ & Set of Spatiotemporal Collocation Points \\
        \hline
        15 & $\mathcal{S}_i$ & Finite Element Subsets of the Spatiotemporal Domain \\ 
        \hline
        16 & $\Xi$ & Standard Finite Element Spatiotemporal Domain \\ 
        \hline
        17 & $\mathcal{R}$ & Region of Attraction (ROA) \\ 
        \hline
        18 & $\mathcal{R}_{x_e}$ & Region of Attraction for Stable Equilibrium Point $x_e$ \\ 
        \hline
        19 & $\mathcal{R}^{\star}$ & ANN Region of Attraction Approximation \\ 
        \hline
        20 & $\mathcal{G}$ & Set of Linear Finite Element Basis Functions \\
        \hline
        21 & $\partial \mathcal{L}_{c}$ & c-Level Set \\
        \hline
        22 & $\mathcal{L}_{c}$ & c-sublevel Set \\
        \hline
        23 & $T$ & Translation Matrix \\ 
        \hline
        24 & $S$ & Scaling Matrix \\ 
        \hline
        25 & $H$ & Transformation Matrix \\ 
        \hline
        26 & $L_{IC}$ & Initial Condition Loss \\
        \hline
        27 & $L_{BC}$ & Boundary Condition Loss \\
        \hline
        28 & $L_{mon}$ & Monotonicity Loss \\
        \hline
        29 & $L_{r}$ & Residual Loss \\
        \hline
        30 & $L_{v}$ & Variational Loss \\
        \hline
        31 & $L_{reg}$ & Regularization Loss \\
        \hline
        32 & $L$ & Total Loss \\
        \hline
        33 & $\phi$ & Viscosity Solution to Yuan-Li HJ PDE \\ 
        \hline
        34 & $\phi^{\star}$ & ANN Viscosity Solution to Yuan-Li HJ PDE Approximation \\
        \hline
        35 & $\psi$ & Dynamical System Flow Trajectory \\
        \hline
        36 & $f$ & Dynamical System Flow \\ 
        \hline
        37 & $g$ & Linear Finite Element Basis Function \\ 
        \hline
        38 & $u$ & Lyapunov Function \\
        \hline
        39 & $u_{x_e}$ & Lyapunov Function for Equilibrium Point $x_e$ \\
        \hline
        40 & $u^{\star}$ & ANN Lyapunov Function Approximation \\
        \hline
        41 & $r$ & PDE Residual \\
        \hline
        42 & $v$ & PDE Variation \\
        \hline
        43 & $v_{g}$ & PDE Variation for Support Function $g$ \\
        \hline
        44 & $\omega_i$ & $i$th Layer Weights Column Vector \\
        \hline
    \end{tabular}
\end{table}

\subsection{Numerical Methods}
\label{SEC: Numerical Methods}

\subsubsection{Fifth Order Weighted Essentially Non-Oscillatory (WEN05) Method}
\label{SEC: Fifth Order Weighted Essentially Non-Oscillatory (WEN05) Method}
Let $\X \subset \R^{d}$ and $\T \subset \R$.
Consider a first order autonomous PDE of the form $0 = h \left( \frac{\partial \phi}{\partial x_1}, \dots, \frac{\partial \phi}{\partial x_d}, \frac{\partial \phi}{\partial t} \right)$ where $\phi(x, t): \X \bigtimes \T \to \R$ is a dynamical variable.
Let $\phi_{i_1, \dots, i_d} = \phi(x_{1, i_1}, \dots, x_{d, i_d}, t)$.
Further, let 

\begin{align}
    \Delta_{k, j}^{+} \phi_{i_1, \dots, i_d} &= \phi_{i_1, \dots, i_{k - 1}, i_{k} + j, i_{k + 1}, \dots, i_d} - \phi_{i_1, \dots, i_d}, \\
    \Delta_{k, j}^{-} \phi_{i_1, \dots, i_d} &= \phi_{i_1, \dots, i_d} - \phi_{i_1, \dots, i_{k - 1}, i_{k} - j, i_{k + 1}, \dots, i_d}.
\end{align}

\noindent Finally, let $\phi_{k, (i_1, \dots, i_d)}^{\pm} = \frac{\partial \phi_{i_1, \dots, i_d}^{\pm}}{\partial x_k}$ denote the spatial derivative terms.
Then the WENO5 spatial derivative approximations are

\begin{align}
    \phi_{k, (i_1, \dots, i_d)}^{\pm} = \frac{1}{12} \left( - \frac{\Delta_{k, -2}^{+} \phi_{i_1, \dots, i_d}}{\Delta x_k} + 7 \frac{\Delta_{k, -1}^{+} \phi_{i_1, \dots, i_d}}{\Delta x_k} + 7 \frac{\Delta_{k, 0}^{+} \phi_{i_1, \dots, i_d}}{\Delta x_k} - \frac{\Delta_{k, 1}^{+} \phi_{i_1, \dots, i_d}}{\Delta x_k} \right) \pm \Phi^{WENO}( a_{k}^{\pm}, b_{k}^{\pm}, c_{k}^{\pm}, d_{k}^{\pm} )
\end{align}

\noindent where 

\begin{align}
    a_k^{\pm} = \frac{\Delta_{k, \pm 2}^- \Delta_{k, \pm 2}^+ \phi_{i_1, \dots, i_d}}{\Delta x_k}, \,\,\, b_k^{\pm} = \frac{\Delta_{k, \pm 1}^- \Delta_{k, \pm 1}^+ \phi_{i_1, \dots, i_d}}{\Delta x_k}, \,\,\, c_k^{\pm} = \frac{\Delta_{k, 0}^- \Delta_{k, 0}^+ \phi_{i_1, \dots, i_d}}{\Delta x_k}, \,\,\, \& \,\,\, d_k^{\pm} = \frac{\Delta_{k, \mp 1}^- \Delta_{k, \mp 1}^+ \phi_{i_1, \dots, i_d}}{\Delta x_k};
\end{align}

\begin{align}
    \Phi^{WENO}(a, b, c, d) = \frac{1}{3} \omega_0 (a - 2b + c) + \frac{1}{6} \left( \omega_2 - \frac{1}{2} \right)(b - 2c + d);
\end{align}

\begin{align}
    \omega_0 = \frac{\alpha_0}{\alpha_0 + \alpha_1 + \alpha_2} \,\,\, \& \,\,\, \omega_2 = \frac{\alpha_2}{\alpha_0 + \alpha_1 + \alpha_2};
\end{align}

\begin{align}
    \alpha_0 = \frac{1}{(\epsilon + S_0)^2}, \,\,\, \alpha_1 = \frac{6}{(\epsilon + S_1)^2}, \,\,\, \& \,\,\, \alpha_2 = \frac{3}{(\epsilon + S_2)^2}; \,\,\, \&
\end{align}

\begin{align}
    S_0 = 13(a - b)^2 + 3(a - 3b)^2, \,\,\, S_1 = 13(b - c)^2 + 3(b + c)^2, \,\,\, \& \,\,\, S_2 = 13(c - d)^2 + 3(3c - d)^2
\end{align}

The WENO5 spatial discretization scheme takes a PDE and expresses it as a system of ODEs with only the temporal variable remaining.
This spatial discretization method is especially suited for PDEs with discontinuities.
See \cite{liu_weighted_1994} for a more detailed look at WENO5 and related methods.

\subsubsection{Third Order Total Variational Diminishing Runge-Kutta (TVDRK3) Method}
\label{SEC: Third Order Total Variational Diminishing Runge-Kutta (TVDRK3) Method}
Let $\T \subset \R$.
Consider an autonomous dynamical system $\dot{\phi} = h(\phi)$ with $\phi: \T \to \R$ and initial condition $\phi_0 = \phi(0)$.
Let $\phi_{i} = \phi(t_i) = \phi(t + i \Delta t)$ where $\Delta t \in \R_{+}$ is the integration time step.
Finally, let

\begin{align}
    \phi^{(1)} &= \phi^{(0)} + \Delta t h \left( \phi^{(0)} \right), \\
    \phi^{(2)} &= \phi^{(1)} + \frac{\Delta t}{4} \left( -3 h \left( \phi^{(0)} \right) + h \left( \phi^{(1)} \right) \right), \\
    \phi^{(3)} &= \phi^{(2)} + \frac{\Delta t}{12} \left( - h \left( \phi^{(0)} \right) - h \left( \phi^{(1)} \right) + 8 h( \phi^{(2)} \right).
\end{align}

\noindent Then the TVDRK3 method approximates $\phi_{i + 1}$ from $\phi_{i}$ by

\begin{align}
    \phi^{(0)} &= \phi_{i}, \\
    \phi_{i + 1} &= \phi^{(3)}.
\end{align}

A temporal integration technique such as TVDRK3 can be applied after a spatial discretization technique such as WENO5 to solve the resulting system of ordinary differential equations.
For more detail on TVDRK3 and related techniques, see \cite{gottlieb_total_1998}.

\end{document}